\documentclass[twocolumn, switch]{article} 

\usepackage{preprint}

\usepackage{graphicx}%
\usepackage{multirow}%
\usepackage{amsmath,amssymb,amsfonts}%
\usepackage{amsthm}%
\usepackage{mathrsfs}%
\usepackage[title]{appendix}%
\usepackage{textcomp}%
\usepackage{manyfoot}%
\usepackage{booktabs}%
\usepackage{algorithm}%
\usepackage{algorithmicx}%
\usepackage{algpseudocode}%
\usepackage{listings}%

\usepackage{siunitx}
\usepackage{tikz}

\DeclareSIUnit\px{px}

\usepackage[numbers,square]{natbib}
\usepackage[utf8]{inputenc}	
\usepackage[T1]{fontenc}	
\usepackage{xcolor}		
\usepackage[colorlinks = true,
            linkcolor = purple,
            urlcolor  = blue,
            citecolor = cyan,
            anchorcolor = black]{hyperref}	
\usepackage{booktabs} 		
\usepackage{nicefrac}		
\usepackage{microtype}		
\usepackage{lineno}		
\usepackage{float}			

\usepackage{lipsum}		

\usepackage{newfloat}
\DeclareFloatingEnvironment[name={Supplementary Figure}]{suppfigure}
\usepackage{sidecap}
\sidecaptionvpos{figure}{c}

\usepackage{titlesec}
\titlespacing\section{0pt}{12pt plus 3pt minus 3pt}{1pt plus 1pt minus 1pt}
\titlespacing\subsection{0pt}{10pt plus 3pt minus 3pt}{1pt plus 1pt minus 1pt}
\titlespacing\subsubsection{0pt}{8pt plus 3pt minus 3pt}{1pt plus 1pt minus 1pt}

\usepackage{tikz,xcolor,hyperref}

\definecolor{lime}{HTML}{A6CE39}
\DeclareRobustCommand{\orcidicon}{
	\begin{tikzpicture}
	\draw[lime, fill=lime] (0,0) 
	circle [radius=0.16] 
	node[white] {{\fontfamily{qag}\selectfont \tiny ID}};
	\draw[white, fill=white] (-0.0625,0.095) 
	circle [radius=0.007];
	\end{tikzpicture}
	\hspace{-2mm}
}
\foreach \x in {A, ..., Z}{\expandafter\xdef\csname orcid\x\endcsname{\noexpand\href{https://orcid.org/\csname orcidauthor\x\endcsname}
			{\noexpand\orcidicon}}
}

\newcolumntype{R}[2]{%
    >{\adjustbox{angle=#1,lap=\width-(#2)}\bgroup}%
    l%
    <{\egroup}%
}

\def\secref#1{Sec.~\ref{#1}}
\def\figref#1{Fig.~\ref{#1}}
\def\tabref#1{Tab.~\ref{#1}}
\def\eqref#1{Eq.~(\ref{#1})}

\renewcommand{\d}[1]{{\mbox{\boldmath$#1$}}}
\newcommand{\m}[1]{{\mbox{{\fontencoding{T1}\sffamily\slshape{#1\/}}}}}

\newcommand{\ical}[1]{{\mbox{\usefont{OT1}{pzc}{m}{it}{#1}}}}

\title{Data-driven Crop Growth Simulation on Time-varying Generated Images using Multi-conditional Generative Adversarial Networks}

\usepackage{authblk}

\author[1\thanks{\tt{ldrees@uni-bonn.de}}]{Lukas Drees\orcidA{}}
\author[2]{Dereje T. Demie\orcidB{}}
\author[3]{Madhuri R. Paul\orcidC{}}
\author[1]{Johannes Leonhardt\orcidD{}}
\author[2]{Sabine J. Seidel\orcidE{}}
\author[3]{Thomas F. Döring\orcidF{}}
\author[4,1]{Ribana Roscher\orcidG{}}

\affil[1]{Remote Sensing Group, Institute of Geodesy and Geoinformation, University of Bonn, Germany}

\affil[2]{Crop Science Group, Institute of Crop Science and Resource Conservation, University of Bonn, Germany}

\affil[3]{Agroecology and Organic Farming Group, Institute of Crop Science and Resource Conservation, University of Bonn, Germany}

\affil[4]{Data Science For Crop Systems, Forschungszentrum Jülich GmbH, Germany}

\begin{document}

\twocolumn[ 
  \begin{@twocolumnfalse} 
  
\maketitle

\begin{abstract}
\textbf{Background:}
Image-based crop growth modeling can substantially contribute to precision agriculture by revealing spatial crop development over time, which allows an early and location-specific estimation of relevant future plant traits, such as leaf area or biomass.
A prerequisite for realistic and sharp crop image generation is the integration of multiple growth-influencing conditions in a model, such as an image of an initial growth stage, the associated growth time, and further information about the field treatment.
While image-based models provide more flexibility for crop growth modeling than process-based models, there is still a significant research gap in the comprehensive integration of various growth-influencing conditions. 
Further exploration and investigation are needed to address this gap.

\textbf{Methods:}
We present a two-stage framework consisting first of an image prediction model and second of a growth estimation model, which both are independently trained.
The image prediction model is a conditional Wasserstein generative adversarial network (CWGAN). 
In the generator of this model, conditional batch normalization (CBN) is used to integrate different conditions along with the input image. This allows the model to generate time-varying artificial images dependent on multiple influencing factors of different kinds.
These images are used by the second part of the framework for plant phenotyping by deriving plant-specific traits and comparing them with those of non-artificial (real) reference images.
In addition, image quality is evaluated using multi-scale structural similarity (MS-SSIM), learned perceptual image patch similarity (LPIPS), and Fréchet inception distance (FID).
During inference, the framework allows image prediction for any combination of conditions used in training; we call this prediction data-driven crop growth simulation.

\textbf{Results:} 
Experiments are performed on three datasets of different complexity. 
These datasets include the laboratory plant \textit{Arabidopsis thaliana} (Arabidopsis) and crops grown under real field conditions, namely cauliflower (GrowliFlower) and crop mixtures consisting of faba bean and spring wheat (MixedCrop).
In all cases, the framework allows realistic, sharp image predictions with a slight loss of quality from short-term to long-term predictions.
For MixedCrop grown under varying treatments (different cultivars, sowing densities), the results show that adding these conditions increases the prediction quality as measured by the estimated biomass.
Simulations of varying growth-influencing conditions performed with the trained framework provide valuable insights into how such factors relate to crop appearances, which is particularly useful in complex, less explored crop mixture systems.
Further results show that adding process-based simulated biomass as a condition increases the accuracy of the derived phenotypic traits from the predicted images. 
This demonstrates the potential of our framework to serve as an interface between an image- and process-based crop growth model.

\textbf{Conclusion:}
The image-based prediction of future plant appearances is adequately feasible by multi-conditional CWGAN. 
The presented framework complements process-based models and overcomes limitations, such as the reliance on assumptions and the low exact field-localization specificity, by realistic visualizations of the spatial crop development that directly lead to a high explainability of the model prediction.
\end{abstract}
\vspace{0.35cm}

  \end{@twocolumnfalse} 
] 


\section{Background}\label{sec:background}

\begin{figure*}[t]
    \centering
    \includegraphics[width=1.0\textwidth]{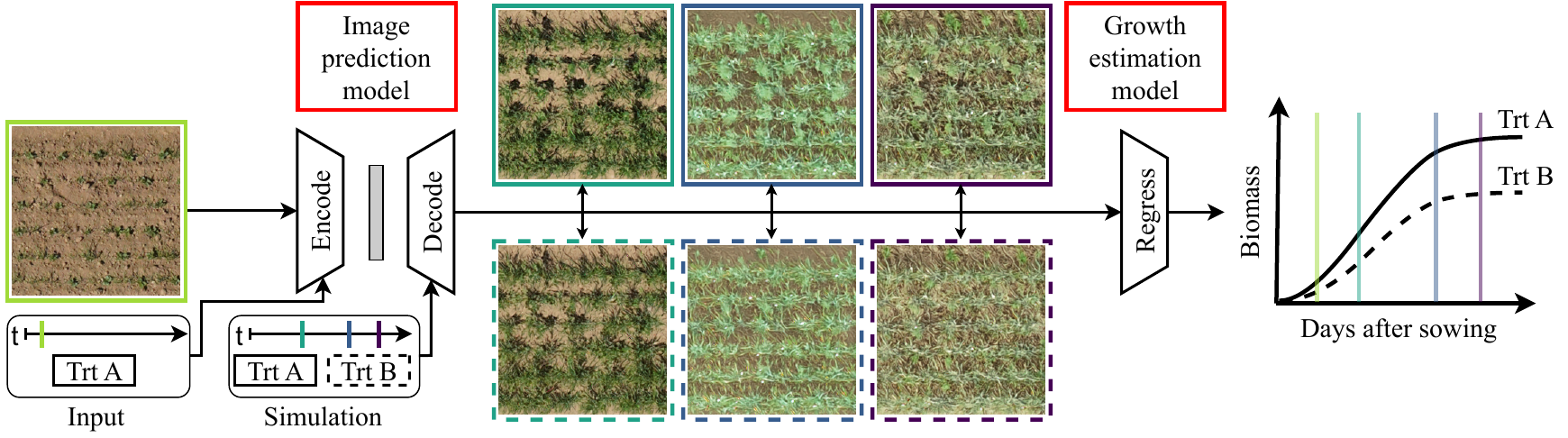}
    \caption{Proposed two-step crop growth simulation framework: In the first step of image prediction, an input image is initially encoded with its associated time (t) and treatment (trt). Then, this encoded representation can be decoded into newly generated images with varying growth stages for different simulation times and treatments. In the second step of growth estimation, target parameters such as projected leaf area or biomass are estimated from the images and analyzed over time. Both models are trained independently.}
    \label{fig:framework}
\end{figure*}

Growing agricultural crops sustainably \footnotetext[1]{$^*$correspondence: \texttt{ldrees@uni-bonn.de}}, i.e., producing sufficient output with high resource use efficiency and minimal negative impacts on ecosystems, requires complex optimization of crop management \cite{shah2019soil}. 
Decisions on the operations during the season include the timing and amounts of fertilization, irrigation, protection against pests and pathogens, weeding, the application of growth regulations, and other activities.
The optimality of most of these operations and their combinations depend on the phenology of crops, i.e., the growth stages and size of the plants. Complex and multiple interactions typically occur between different management factors, crop genotypes, and variable environmental factors, affecting crop performance differently at different growth stages. 
Because of this complexity, identifying optimized crop management is not trivial, and various ways have been developed to tackle this problem and to understand crop responses to complex management~$\times$~environment interactions. 
Two complementary approaches are experimental field trials and process-based (mechanistic) crop growth monitoring. 
While field experimentation integrates actual environmental and management conditions, it is limited in time and space and can only test a low number of such conditions. 
Crop growth modeling, on the other hand, while allowing the simulation of multiple conditions, including future environments, is always a simplification of the situation in the field and may be limited in predicting realistic responses of crops, especially under a changing climate \cite{sarkar2020low}. 
Because of the central role of crop phenology in agronomic decision-making, it is useful to be able to predict future crop growth stages and crop appearance in the season. 
One pathway towards this goal is the automatic generation of crop images derived from images taken during earlier stages.
This is particularly difficult but also useful in crop mixtures, where interactions occur between two or more crop species grown together on the same field.

As an example, cereal and legume crop mixtures are known to improve resource use efficiency \cite{jensen2010faba}, enhance nutrient acquisition \cite{li2007diversity}, maximize system productivity through complementarity, especially on low input land limited by nitrogen deficiency \cite{peoples2009contributions}, and reduce weeds, diseases, and insect pest infestations \cite{bedoussac2015ecological}.
Nevertheless, many farmers do not consider crop mixtures as an option, often due to a knowledge gap in species, cultivar, and treatment selection, which results in performance uncertainty \cite{yu2016meta}. 
One approach to overcome this uncertainty and to deal with complexity is the use of predictive crop modeling.

The differences between predictive crop growth models are manifold. Process-based models are based on biological and physical relationships and aim to represent the mechanics of plant growth and thus have a high interpretability.
They are also suitable for long-term predictions and can be generalized to different locations, but both require a complex calibration to the respective environment.
Image-based crop growth models, on the other hand, are data-driven, with information on the actual crop environment encoded in the image.
By using machine learning to process this image data, data-driven models can build complex relationships \cite{tsaftaris2016machine} without relying on simplified assumptions, like process-based models.
This makes the modeling process less interpretable, but the result that comprises a predicted image, showing a realistic future spatial plant development, and derived phenotypic traits can be explained in a better way as it is human-understandable.
The predicted image is highly versatile, which is particularly interesting for crop mixtures, e.g., to count the future number of crops at certain field positions or to visualize, and thus better understand, how two species behave and compete with respect to certain influencing factors.
Therefore, this work aims to extract more insight from image-based models to complement missing facets of existing well-established crop growth models.

In recent years, the most widely used method for image generation in plant science has been Generative Adversarial Networks (GANs) \cite{goodfellow2014generative}, as they have proven to generate high-quality images.
In particular, its variant conditional GAN (CGAN) has found a wide application, e.g., to generate realistic plant images \cite{nazki2020unsupervised,madsen2019generating,zhu2018data} for the purpose of data augmentation, or segmentation \cite{kierdorf2021behind}.
While these works operate in the same temporal domain, few works exist that incorporate the factor of time to generate and analyze probable future growth stages.
Yasrab et al. \cite{yasrab2021predicting} generate segmentation images of future root and shoot systems of Arabidopsis (\textit{A. thaliana}) and Komatsuna (\textit{Brassica rapa}) based on a time series of past images.
However, their GAN model is limited to observation times with fixed constant intervals, which severely limits the space of possible input time series and makes long-term predictions difficult.
Furthermore, due to significant differences in the bit depth, the generation of segmentations is much less complex than the generation of artificial plant images, which can be considered as artificial sensor data.
Drees et al. \cite{drees2021pix2pixbrassica} show long-term predictions of realistic images of the above-ground plant phenotype, although it has the disadvantage that time is not explicitly included as a condition so that the image prediction is limited to predefined growth prediction steps between fixed growth stages.
This challenge can be addressed by extending the generator with modules responsible for integrating the time factor, such as a combination of positional encoded time points and a transformer encoder, as shown in \cite{drees2022time}.
This allows the flexible integration of multiple time points as a condition, as well as the generation of an arbitrary growth stage in the output.
However, the image quality is not optimal because the model is limited to a small bottleneck dimension due to a parameter-intensive Transformer encoder.
Further, the evaluation in this work is based only on classical metrics, such as structural similarity, but lacks crucial plant-specific evaluations that demonstrate actual usability by deriving phenotypic traits.
In general, all the aforementioned methods have the disadvantage that plant growth is greatly simplified by considering only other growth stages, so the time factor in the input, while in fact, it is subject to complex mechanisms.
Miranda et al. \cite{miranda2022controlled} attempt to get closer to this complexity by integrating more conditions into the growth modeling, which allows them to generate controlled and more explainable output images.
However, the method is limited to continuous conditions and to a predefined growth prediction step from a fixed early growth stage to a fixed later growth stage, which is unfavorable in agricultural practice. 
In general, integrating multiple conditions is a non-trivial task, as conditioned image generation tends to generate deterministic and less diverse outputs up to mode collapse \cite{shahbazi2022collapse}. 
There are many different ways of integrating conditions from concatenation \cite{mirza2014conditional} over auxiliary classifiers \cite{odena2017conditional} and latent projection \cite{karras2020training} to conditional batch normalization \cite{dumoulin2016learned,brock2018large}.
This work uses the latter because it allows the intuitive integration of multiple conditions while maintaining the stochasticity of the model to create an adequate distribution of generated plants.


An overview of our growth simulation framework is depicted in \figref{fig:framework}.
It is a two-step procedure in which time-varying images are first generated with the image prediction model and then analyzed with an independently trained growth estimation model.
An important novelty in the image prediction model, which is a Conditional Wasserstein GAN (CWGAN), is the integration of multiple conditions of different types, that is, images (2D spatial continuous variables), time points (discrete), treatment information (categorical), and daily simulated biomass (continuous).
Since the biomass is process-based simulated, we demonstrate that the image prediction model can serve as an interface that makes the output of process-based crop growth models more explainable by visualizing the spatial crop development.
Conditioning is realized by conditional batch normalization in both parts, the encoder and decoder, of the CWGAN generator.
This enables simulations during inference, i.e., while fixing initial conditions (input image, time, and treatment), for other growth stages, conditions can be varied -as required- to generate multiple realistic predictions, as shown in \figref{fig:framework}.
Experiments have been conducted on different datasets of varying complexity, from the laboratory plant \textit{Arabidopsis thaliana} (Arabidopsis) to real field data with cauliflower (GrowliFlower) and crop mixtures (MixedCrop).
In addition to classical GAN evaluation metrics, we evaluate the quality of generated images through the growth estimation model, which acts as a plant phenotyping module, by comparing (depending on the dataset) either the projected leaf area or the biomass estimated from generated and real images.
For crop mixtures, this allows us to make a comparison between our image-based crop growth simulation and a classic process-based one, which was used to establish the growth estimation model.
A transferability experiment demonstrates that our framework has the potential to be transferred to crop mixtures in another field with different environmental conditions.

\section{Materials and Methods}\label{sec:methods}
This section introduces the data basis (\secref{sec:data}) and the framework\footnote{Source code is publicly available at \url{https://github.com/luked12/crop-growth-cgan}}, where a 2-step approach is followed.
First, an image is predicted (\secref{sec:ip}), and second, the growth is estimated using plant phenotyping (\secref{sec:ge}). 
While existing state-of-the-art models are used for growth estimation, which is fine-tuned on our data, the methodological focus of this work is on the first part, image prediction.
We also provide details about the process-based crop growth model (\secref{sec:simulation}) used to evaluate and analyze image-based predictions of crop mixtures.

For a clear distinction, we call a model output estimation if it has the same time as the input and prediction if there is a time shift $\Delta t=t_{\text{gen}}-t_{\text{in}}$, so positive $\Delta t$ means prediction into the future and negative $\Delta t$ means prediction into the past.

\subsection{Data}\label{sec:data}

\begin{table*}[t]
\centering
\begin{tabular}{lrrrr}
\toprule
& \multicolumn{1}{c}{\multirow{2}{*}{Arabidopsis}} & \multirow{2}{*}{GrowliFlower} & \multicolumn{2}{c}{MixedCrop} \\
& \multicolumn{1}{c}{} & & Mixed-CKA & Mixed-WG \\
\midrule
\# images & 54\,384 & 102\,264 & 21\,371 & 18\,800\\
observation period [\SI{}{d}] & 18 & 71 & 113 & 109 \\
\# times\footnotemark[1] (Cond.: \textbf{t})& 850 & 12  & 11 & 10 \\
\# sequences\footnotemark[2] & 64 & 8\,522 & 2226 & 2212 \\
\# train sequences & 40 & 6\,572 & 1555 & 1580\\
\# val sequences & 8 & 979 & 311 & 316\\
\# test sequences & 16 & 971 & 311 &  316 \\
$\varnothing$ images/sequence & 850 & 12 & 9.60 & 8.50 \\
image size [\SI{}{px}] & 256 & 256 & 256 & 256 \\
GSD [\SI{}{mm}] & 0.23 & 3.10 & 5.67 & 5.67 \\
diff. treatments (Cond.: \textbf{trt})& \texttimes & \texttimes & \checkmark (76) & \checkmark (76) \\
sim. biomass (Cond.: \textbf{bm})& \texttimes & \texttimes & \checkmark & \checkmark \\
\midrule
GEM: \# train images & 512 & 1\,541 & 15\,017 & 13\,154  \\
GEM: \# val images & 148 & 326 & 3\,177 & 2\,823 \\
GEM: \# test images & 148 & 330 & 3\,177 & 2\,823 \\
\bottomrule
\end{tabular}
\footnotetext[1]{The number of different time points equals the max. sequence length; for Arabidopsis, it is clearly greater than the period because up to four images were taken per hour.}
\footnotetext[2]{The number of sequences equals the number of different plants in Arabidopsis and spatially separated field patches in GrowliFlower and MixedCrop.}
\caption{Dataset characteristics. The upper block indicates the image specifications for the image prediction model, where the different conditions time (t), treatment (trt), and biomass (bm) are highlighted, and the bottom block displays the number of images used to train, validate, and test the resp. growth estimation model (GEM), which is trained independently on individual images without sequence information.}
\label{tab:data_stats}
\end{table*}

\begin{figure}[t]
    \centering
    \includegraphics[width=1.0\columnwidth]{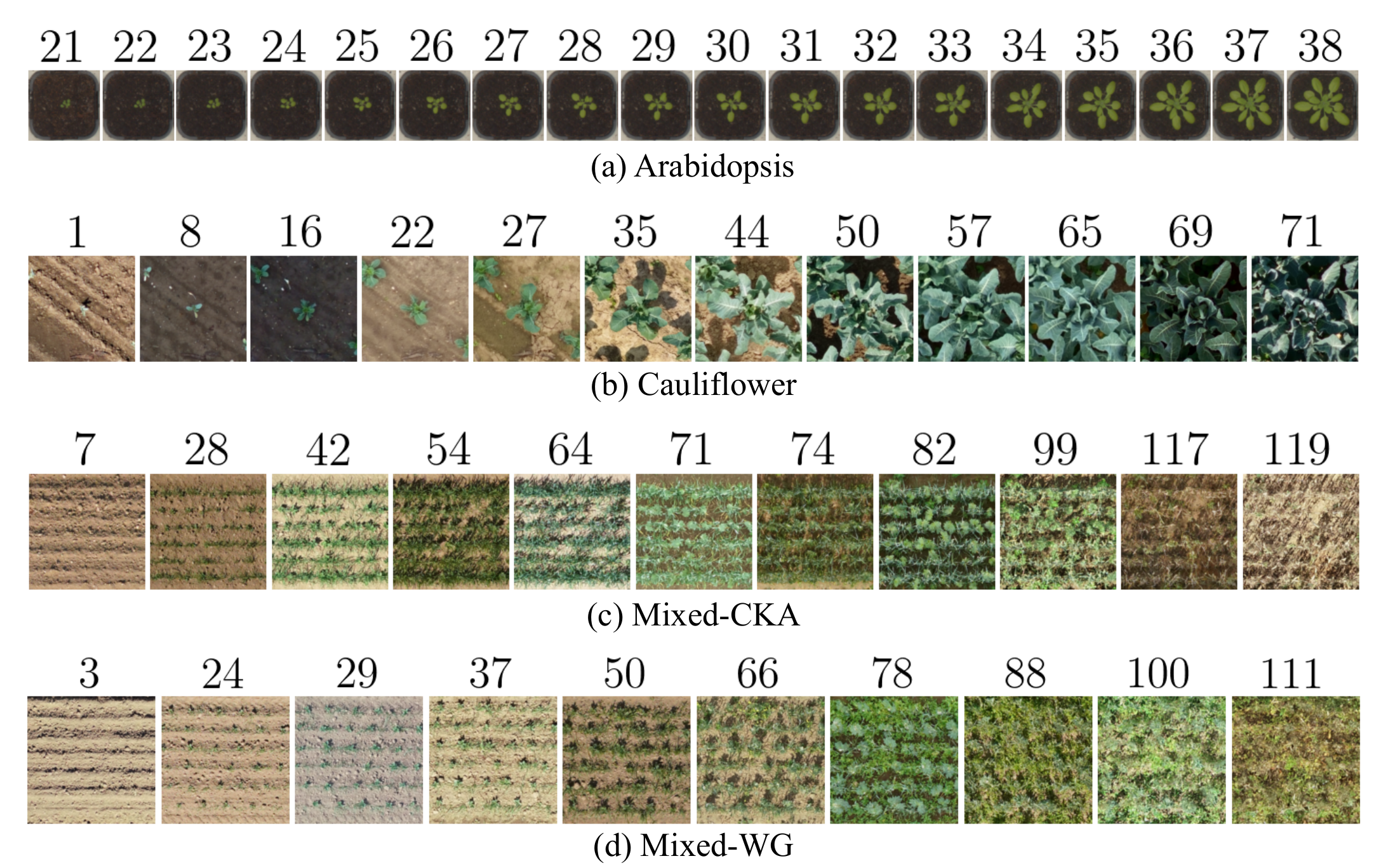}
    \caption{Example evolution over time of one plant resp. from each of the datasets (a) Arabidopsis, (b) GrowliFlower, (c) Mixed-CKA, and (d) Mixed-WG visualized by georeferenced clips from RGB orthophotos. The number above the images indicates the growth stage for (a),(c), and (d) in days after sowing [DAS] and for (b) in days after planting [DAP].}
    \label{fig:data}
\end{figure}

Experiments are set up on three different datasets: Arabidopsis, GrowliFlower, and MixedCrop, all containing RGB-image time series/sequences of plants (\figref{fig:data}).
They meet the minimum requirement of having aligned images, which means that all images of a sequence show the exact same region from the same perspective and resolution over time.
Ideally, lightning conditions are constant, which is only the case for the Arabidopsis dataset.
Beyond that, they differ in essential aspects such as overall size, type of plants, heterogeneity of images, and number, as well as regularity of acquisition times during the vegetation period.
Notably, additional conditions on treatment and daily simulated biomass are available only for MixedCrop, all listed in \tabref{tab:data_stats}.

\textbf{Arabidopsis}.
The Arabidopsis dataset \cite{bell2016aberystwyth} includes 80 different Arabidopsis (\textit{Arabidopsis thaliana}) plants recorded on 4 trays of 20 plants each over a \SI{35}{d} period using an IDS UI-5480SE camera (Tamron \SI{8}{mm} f1.4 lens, \SI{5}{MP}).
The camera was mounted on a robotic arm in a controlled laboratory environment, ensuring image alignment.
All tray images are corrected for barrel distortions with a provided calibration script \cite{bell2016aberystwyth} and then manually cropped at the edges of the pots, resulting in images that have a single plant in the center region.
We focus on images from 18 days of early developmental stages of Arabidopsis thaliana from day 21 after sowing, which is shortly after plant emergence, to day 38 after sowing.
Any plants that were already removed from the experiment before day 38 or that protruded beyond the edge of the pot after day 38 were removed, leaving 64 plants.
Please note that the number of images per sequence clearly exceeds the duration of the observation period in [d] because not only one image per day was taken, but up to four per hour.

\textbf{GrowliFlower}. 
Cauliflower image sequences from the GrowliFlower dataset \cite{kierdorf2023growliflower} contains a total of 102\,264 images of cauliflower (\textit{Brassica oleracea var. botrytis}) in 2021 from a field in Bornheim, Rhein-Sieg Kreis, Germany.
We use the images in a period of 71 days after planting and exclude images after harvest.
The images are orthophoto crops taken from a drone equipped with a Sony Alpha 7R III camera (Zeiss/Batis 2.0 lens, \SI{47.4}{MP}). 
The geo-referencing of the orthophotos allows aligned plant-centered cropping at the same position at each time point.
However, compared to Arabidopsis, there is not only one plant per image, but more heads are visible at the image edges and overlap to later growth stages.

\textbf{MixedCrop}. 
The MixedCrop data are from a 2020 and 2021 PhenoRob crop mixture experiment described in detail by Paul et al. \cite{paul2023effects}.
Two different cultivars of faba bean (FB, \textit{Vicia faba}) and twelve different entries of spring wheat (SW, \textit{Triticum aestivum}) were sown in mixtures of a 1:1 ratio, which means 50~\% of seeds of each species from the respective monoculture as well as in monocultures. 
The field experiments were conducted at two research sites of the University of Bonn in the Rhein-Sieg-Kreis, Germany, namely, Campus Klein-Altendorf (CKA, near Rheinbach) and at Wiesengut (WG, near Hennef).
Coupled with two different seeding densities i.e. low (L) 80\% and high (H) 120\% of the recommended sole crop densities (400 seeds~$\text{m}^{-2}$ for SW and 45 seeds~$\text{m}^{-2}$ for FB), this results in $(2\cdot12+2+12)\cdot2=76$ different treatments, which were replicated four times, or, in case of the faba bean monocultures, eight times, resulting in a total of 320 different plots of size $\SI{10}{m}\times\SI{1.5}{m}$ at each of the two sites. 
Both experimental sites are located about 30 km apart and have significantly different growing conditions because Mixed-CKA is managed conventionally and Mixed-WG organically. 
The dynamic process-based model used to generate the simulated daily dried biomass values for SW and FB was calibrated and tested against the observation data of the experiment carried out at CKA 2020, 2021, and WG 2020. 
However, the images used in this work are only from CKA 2020 and WG 2020.

The image acquisition was conducted 11 times for Mixed-CKA and 10 times for Mixed-WG by UAV equipped with an FC6310 camera (\SI{1}{''} CMOS \SI{8.8}{mm}, \SI{20}{MP}).
The 320 field plots are positional-aligned cropped from the geo-referenced orthophotos before being horizontally rotated and plot-centered cropped into seven non-overlapping and square image clippings. 
Due to orthophoto corruptions and destructive field measurements, some sections were manually removed, resulting in a final number of 21\,371 images for Mixed-CKA and 18\,800 images for Mixed-WG.
For Mixed-WG, a significant spatial alignment error was noticed by visual inspection, which is up to \SI{10}{cm}, but inconsistent across the images and, therefore, difficult to filter out.
Since \SI{10}{cm} corresponds approximately to the spatial extent of a faba bean plant at 20 days after sowing (DAS), the offset is well visible in the early images.
For this reason, Mixed-WG is not used for training; instead, it is intended to check the transferability, i.e., the model learned on Mixed-CKA and applied on Mixed-WG.

In addition, a variety of other data were collected in this crop mixture experiment, including weather, soil, and nutrient parameters as well as height and biomass measurements \cite{paul2023effects, demie2022mixture} that are used in this work to calibrate and evaluate a process-based crop growth model as described in \secref{sec:simulation}. \\

The datasets vary in their complexity: The challenges of the GrowliFlower and MixedCrop datasets are the considerable gaps of different lengths between the recording times.
In addition, there are large spectral differences both within each time series and between Mixed-CKA and Mixed-WG, mainly due to different solar radiations, cloud coverings, and soil moistures during the overflights.
Compared to the other datasets, MixedCrop is the most challenging due to its small size combined with a large number of overlapping mixed crops, even at early growth stages.
All images are resized to a uniform size of $\SI{256}{px} \times \SI{256}{px}$ for the experiments, resulting in different ground sample distances (GSD) from \SI{0.23}{mm} to \SI{5.67}{mm}.
For all experiments, the image sequences are divided into the same spatially separated training, validation, and test sets.

\subsection{Image prediction}\label{sec:ip}
For image prediction, we build a multi-conditional Wasserstein GAN with gradient penalty (CWGAN-GP) \cite{gulrajani2017wgangp} from several state-of-the-art components.
The network consists of a generator $\mathcal{G}_\theta$ and a critic $\mathcal{D}_\delta$, where $\mathcal{G}_\theta$ predicts images and $\mathcal{D}_\delta$ estimates a score for generated and real images.
A special focus is on the integration of multiple conditions of different types in the architecture as described in \secref{sec:ip-network}.

\subsubsection{Conditional Wasserstein GAN objective}
In the generator, a target image $\m{X}_{\text{gen}}=\mathcal{G}_\theta(\m{X}_{\text{in}},\d{y},\d{z})$ is generated from an input image $\m{X}_{\text{in}}$, conditions $\d{y}$ that split into $[\d{y}_\text{in},\d{y}_\text{gen}]$, and noise $\d{z}\sim \mathcal{N}(0,1)$.
Both $\d{y}_\text{in}$ and $\d{y}_\text{gen}$ represent multi-conditioning, which can be composed of several of the following conditions: categorical (class) variables $c$, discrete variables $t$, and continuous variables \d{b}.
In the critic, either the reference $\mathcal{D}_\delta(\m{X}_{\text{ref}},\m{X}_{\text{in}},$\d{y}$)$ or the generated image $\mathcal{D}_\delta(\m{X}_{\text{gen}},\m{X}_{\text{in}},$\d{y}$)$ are presented along with input image and conditions.
The critic estimates a score for both real and generated input, which is capable of enforcing the minimization of the Wasserstein distance between the two distributions.
The objective of adversarial training is to optimize the parameters $\theta$ and $\delta$ by maximizing the objective function $L_{\text{GAN}}(\mathcal{G}_\theta,\mathcal{D}_\delta)$ by $\mathcal{D}_\delta$ and minimizing it by $\mathcal{G}_\theta$.
\begin{equation} \label{eq:obj}
\theta^*, \delta^* = \arg \min_\theta \arg \max_\delta L_{\text{GAN}}(\mathcal{G}_\theta,\mathcal{D}_\delta)
\end{equation}
\eqref{eq:objfunc} represents $L_{\text{GAN}}(\mathcal{G}_\theta,\mathcal{D}_\delta)$ with the classic CWGAN objective in the first line \cite{arjovsky2017wasserstein}, added with the gradient penalty term in the second line to enforce the required 1-Lipschitz continuity of $\mathcal{D}_\delta$ \cite{gulrajani2017wgangp}.
\begin{equation} \label{eq:objfunc}
\begin{aligned}
    L_{\text{GAN}}(\mathcal{G}_\theta, \mathcal{D}_\delta) = 
    &\mathbb{E}_{{\scriptsize(\textbf{\textit{z}},\m{X}_{\text{in}}},{\scriptsize \textbf{\textit{y}})}}
    [\mathcal{D}_\delta(\mathcal{G}_\theta(\m{X}_{\text{in}},\d{y},\d{z}),\m{X}_{\text{in}},\d{y})]\\
    &- \mathbb{E}_{\scriptsize(\m{X}_{\text{ref}},\m{X}_{\text{in}},{\scriptsize \textbf{\textit{y}}})}
    [\mathcal{D}_\delta(\m{X}_{\text{ref}},\m{X}_{\text{in}},\d{y})]\\
    &+ \lambda_{\scriptsize\text{GP}}\mathbb{E}_{\scriptsize(\m{X}_{\text{in}},\hat{\m{X}})}
[(\lVert \nabla_{\scriptsize\hat{\m{X}}}\mathcal{D}_\delta(\m{X}_{\text{in}},\hat{\m{X}})\rVert_{\scriptsize2}-1)^2]
\end{aligned}
\end{equation}
The gradient penalty is computed by blending a generated image with a reference image, resulting in $\hat{\m{X}}=\epsilon\m{X}_{\text{ref}}+(1-\epsilon)\mathcal{G}_\theta(\m{X}_{\text{in}},\d{y},\d{z})$, where $\epsilon$ is a random value in the range [0, 1], and its impact is controlled by $\lambda_{\scriptsize\text{GP}}$. Using $L_{\text{GAN}}(\mathcal{G}_\theta, \mathcal{D}_\delta)$ minimizes the Wasserstein-1 distance, sidestepping issues like mode collapse and vanishing gradients in classic GAN training.

\subsubsection{Network architecture with multi-conditioning}\label{sec:ip-network}
\textbf{Generator}. 
The generator consists of an encoder $\mathcal{P}$ that compresses the input image and conditions related to the input image into a latent representation $\xi=\mathcal{P}(\m{X}_{\text{in}},\d{y}_{\text{in}})$ and a decoder $\mathcal{Q}$ that generates the target image from this latent representation, the conditions for the image to be generated and a stochastic component $\m{X}_{\text{gen}} = \mathcal{Q}(\xi,\d{y}_{\text{gen}},\d{z})$.
While for image encoding a ResNet-18 backbone \cite{he2016deep} without final fully connected layer and global average pooling is used, decoding works architecturally inverse to that.
To integrate the conditions, all batch normalization layers are replaced by conditional batch normalization layers (CBN) \cite{de2017modulating}, where the learnable affine parameters of classical batch normalization layers \cite{ioffe2015batch} are conditioned on some auxiliary variable \d{a}. In our case, \d{a} are embeddings of the conditions \d{y} using an embedding function $\Phi$.
In particular, the encoder's CBN layers are conditioned on the embeddings related to the input image $\d{a}_\text{in} = \Phi(\d{y}_\text{in})$, while the decoder's CBN layers are conditioned on the embeddings related to the image to be generated $\d{a}_\text{gen} = \Phi(\d{y}_\text{gen})$.
Specifically, the embedding function is condition-type-specific since \d{y} can consist of conditions of up to 3 different types, namely discrete temporal information $t$, categorical class information $c$, and continuous variables $\d{b}$. 
So individual embeddings are performed for each type of condition in \d{y}, which are then concatenated to \d{a}.
\begin{equation} \label{eq:multi-c}
\begin{aligned}
&\d{y}_{\text{in}} = [t_{\text{in}},c_{\text{in}},\d{b}_{\text{in}}]\\
&\d{a}_{\text{in}} = [\Phi_t(t_{\text{in}}),\Phi_c(c_{\text{in}}),\Phi_b(\d{b}_{\text{in}})]\\
&\d{y}_{\text{gen}} = [t_{\text{gen}},c_{\text{gen}},\d{b}_{\text{gen}}]\\
&\d{a}_{\text{gen}} = [\Phi_t(t_{\text{gen}}),\Phi_c(c_{\text{gen}}),\Phi_b(\d{b}_{\text{gen}})]
\end{aligned}
\end{equation}
Here, the temporal embedding $\Phi_t$ consists of positional encoding of discrete time points followed by a two-layer MLP with a sigmoid linear unit (SiLU) function in between.
The class embedding $\Phi_c$ represents a classic lookup table embedding that maps indices of categorical class variables to a continuous vector representation.
In order to embed a vector of continuous values in $\Phi_b$, a two-layer MLP with SiLU function in between is used.
In the experiments, the conditions $c$ and \d{b} are not always used, then embedding and resp. concatenating of unused conditions is omitted.
Notably, $t_{\text{in}}$/$t_{\text{gen}}$ and $c_{\text{in}}$/$c_{\text{gen}}$ are scalars representing in this work time ($t$) and treatment ($c$), respectively, while $\d{b}_{\text{in}}$/$\d{b}_{\text{gen}}$ are vectors, and in this work are 2-dimensional due to SW and FB biomass. 
However, after embedding the individual components of \d{y}, it is ensured that $\Phi_t(\d{t})$, $\Phi_c(\d{c})$, and $\Phi_a(\d{b})$ all represent continuous vectors of the same 64-dimensional embedding size, which avoid prior weighting of different conditions.
Besides, CBN has already included a linear embedding for all conditions, but the additional condition-type-specific embedding has stabilized the training process.

To also incorporate stochasticity into the network, a random 128-dim noise vector $\d{z}\sim \mathcal{N}(0,1) \in \mathcal{Z}$ is generated and via noise mapping network $\ical{f}:\ical{Z} \mapsto \ical{W}$ inspired by StyleGAN \cite{karras2019style} projected to the latent code $\d{w} \in \mathcal{W}$, that matches the channel dimension of the latent representation $\xi$.
The mapping network $\ical{f}$ is a shallow three-layer linear embedding network, which gradually projects to 128-dimensional \d{z} to the 512-dimensional \d{w}, which corresponds to the channel size of the ResNet-18 latent representation.
After repeating \d{w} for the spatial dimension (global average pooling is omitted), it is finally added to $\xi$.\\

\textbf{Critic}.
The critic takes either the generated $\m{X}_{\text{gen}}$ or reference image $\m{X}_{\text{ref}}$ along with the input image $\m{X}_{\text{in}}$, and the conditions $\d{y}$ as input.
The images are concatenated channel-wise in the input and initially passed through a convolutional layer and LeakyReLU activation.
This is followed by several convolutional blocks consisting of a convolutional layer, instance normalization, and LeakyReLU up to a spatial dimension of $[16\times16]$.
Since batch normalization should be avoided in the Wasserstein critic \cite{gulrajani2017wgangp}, the conditions are not integrated in this case with CBN.
Instead, each condition is first embedded to dimension 256 with a different embedding function $\Psi$ than $\Phi$ in the generator, but the architecture of the embedding functions inside $\Psi$ and $\Phi$ are the same.
Then, embedded conditions are reshaped, and channel-wise concatenated to the intermediate critic representation of spatial size $[16\times16]$.
Note that here, the conditions of both the input image and the image to be generated are concatenated.
From this concatenated representation, the final score is generated with further convolutional blocks.
Previous experiments have shown that the training converges significantly better with an intermediate fusion of the conditions than with a fusion directly in the critic input.

\subsubsection{Optimization and hyperparameter}
The data sampling is special since, due to the temporal condition, multiple reference images can be used for every input image. 
Thus, in each epoch, we iterate over all training images, which are then used as input images. 
For each input image, a random image of the same plant is sampled, representing the reference plant.
The conditions $\d{y}_{\text{in}}$ and $\d{y}_{\text{gen}}$ are drawn according to the sampled images.
This causes that during the training $c_{\text{in}}$=$c_{\text{gen}}$ because the treatment class does not change over time.
To calculate test scores, the sampling procedure is identical, i.e., each test image represents an input image once and gets assigned a random growth stage as the reference image to be generated.
For inference, the conditions can be varied arbitrarily, what we call data-driven simulation.
So a treatment change $c_{\text{in}}\neq c_{\text{gen}}$ is possible, \d{b} does not have to fit the reference values, and $t$ can deviate from the training range.

Adam optimizer is used with a learning rate of 1e-4 for both $\mathcal{G}_\theta$ and $\mathcal{D}_\delta$ optimization.
Regardless of the number of conditions, the models are trained for 5000 epochs, after which the best epoch is selected based on the lowest LPIPS on the validation data.
As image augmentations, horizontal and vertical flipping, \SI{90}{\degree} rotations, slight translations within a random affine transformation, and ShadowOut, which is a semi-transparent version of CutOut \cite{devries2017improved}, are applied simultaneously to input and reference or generated image.
Using a single NVIDIA A100-PCIE-40GB and a batch size of 64, the training duration is between \SI{13}{d} and \SI{35}{d}, depending on the dataset size.

\subsubsection{Evaluation of image quality}
To evaluate the quality of the generated images, we use a well-established set of GAN evaluation metrics.
For the direct comparison between generated and reference images of the same time point, we use the Multi-scale Structural Similarity Index Measure (MS-SSIM \cite{wang2003msssim}, optimal: 1) and the Learned Perceptual Image Patch Similarity (LPIPS \cite{zhang2018unreasonable}, optimal: 0).
While MS-SSIM compares the generated with the reference image directly at different resolutions of the image space, LPIPS evaluates the similarity of image patch activations in the VGG-embedded latent space, which has been shown to have a high correlation with human perception.
In addition, the Fréchet Inception Distance (\text{FID} \cite{heusel2017gans}, optimal: 0) is used to compare not only the quality but also the diversity of the generated image distribution with the real image distribution of the test dataset.
However, for long-term predictions far into the future or past, that means a large difference exists in the growth stage of the input image and the image to be generated, it is not expected that generated and reference images match at the pixel level. 
Although FID will degrade less as long as the plants fit into the distribution of each growth stage, poor results are to be expected for MS-SSIM and LPIPS in such cases.
In order to evaluate whether useful plant-related traits can still be derived, we use growth estimation models, which determine leaf area (\secref{sec:eval_pla}) and biomass (\secref{sec:eval_bm}) from the generated images.

\subsection{Growth estimation}\label{sec:ge}
The part of growth estimation is realized, depending on the dataset and plant type, either by instance segmentation to estimate projected leaf area or by image regression to estimate biomass.
Both can also be considered plant phenotyping based on state-of-the-art neural networks. 

\subsubsection{Estimation of projected leaf area}\label{sec:eval_pla}
For Arabidopsis and GrowliFlower, growth is determined using the plant trait projected leaf area (PLA).
Both datasets are well suited for this purpose because different plants do not overlap until advanced growth stages.
The PLA is derived as an image-wise pixel sum of plant segmentations predicted with a Mask R-CNN instance segmentation model \cite{he2017mask}.
For this, two models, with pre-trained ImageNet weights \cite{krizhevsky2012imagenet}, are fine-tuned on a few images of the respective plant dataset, for which reference segmentation masks are available.
By multiplying the PLA with the squared dataset-dependent ground sample distance (GSD), we report PLA in the unit \SI{}{mm^2} for Arabidopsis and \SI{}{cm^2} for GrowliFlower, or for comparability normalized in units of \SI{}{\%/image}, which is achieved by dividing the PLA by the image size.
In this work, PLA is not calculated for the whole image but only out of the segmentation predictions for the center plant, which is especially relevant for GrowliFlower, where there are, in most cases, multiple plants per image.
To compare the PLA of a single generated and reference image pair, we use $\Delta\text{PLA}=\text{PLA}^{\text{gen}}-\text{PLA}^{\text{ref}}$.
For MixedCrop, PLA cannot be extracted with sufficient accuracy at the pixel level for the individual crop species due to the fine structure of the wheat ears, enormous plant overlap, and a lack of annotated images \cite{marashdeh2022semantic}.
The accuracy evaluation of the trained instance segmentation models can be found in \secref{sec:acc_pla}.

\subsubsection{Estimation of biomass}\label{sec:eval_bm}
Instead of PLA, for MixedCrop, dried biomass (BM) in tons per hectare [\SI{}{t/ha}] is to be derived from the images as a growth indicator, divided into the two mixture species spring wheat (SW) and faba bean (FB).
To estimate both with one model, a ResNet-18 \cite{he2016deep} is used, modifying the last linear layer to two output neurons, which are activated with ReLU, since only positive biomass values are possible.
The mean squared error (MSE) function is used as the loss function.
We use weights from a pre-training with ImageNet \cite{krizhevsky2012imagenet} and fine-tune on MixedCrop images and corresponding reference biomass values.
These reference biomass values are not actual in-field measurements but come from a process-based crop growth model for mixtures (see \secref{sec:simulation}) that provides simulated SW and FB biomasses dynamically for each image time point.
Notably, we use the same simulated biomass values that are used as conditions in the image prediction part of the framework.
However, this dual use is methodologically not critical since the image prediction part and the growth estimation part are trained independently of each other.
Similar to PLA, we use $\Delta\text{BM}=\text{BM}^{\text{gen}}-\text{BM}^{\text{ref}}$ to report biomass deviations between two images.
Overall, estimating biomass from bird's eye view imagery has three main challenges and inherent sources of error.
First, biomass is a 3D quantity derived from 2D images. 
Second, the process-based crop growth model only estimates dried biomass for all growth stages, which is used as a reference for training the growth estimation model. 
However, the images show plants with their actual humidity (fresh matter), which changes over time.
Third, the simulation result varies only treatment-wise, but it is likely that plants of the same treatment will develop differently in multiple replications in the field due to different soil conditions.
For the discussion about the biomass estimation results and accuracies, see \secref{sec:acc_biomass}.

In the evaluation for a whole test set with $N$ images, the mean absolute error (MAE) and the mean error (ME) are calculated as follows between plant traits (PT) of the generated and the reference image, whereby either PLA or BM serve as PT.
\begin{equation}
\text{MAE}=\frac{\sum_{i=1}^N\lvert\text{PT}^{\text{gen}}_i-\text{PT}^{\text{ref}}_i\rvert}{N}
\end{equation}
\begin{equation}
\text{ME}=\frac{\sum_{i=1}^N\text{PT}^{\text{gen}}_i-\text{PT}^{\text{ref}}_i}{N}
\end{equation}
Here, the quantity measure ME indicates whether the PT is overall underestimated (ME negative) or overestimated (ME positive).
For whole agricultural fields, the mean error (ME) is informative, in case it is not as important to accurately determine the yield of individual field regions but rather to evaluate whether the overall mean predictive error for the entire field is low.

\subsubsection{Process-based modeling of crop mixtures}\label{sec:simulation}
The process-based crop growth simulations were conducted in the modeling platform SIMPLACE (Scientific Impact Assessment and Modeling Platform for Advanced Crop Ecosystem Management) \cite{enders2023simplace}.
Different SimComponents (submodels) in the SIMPLACE framework were combined, namely LINTULPhenology, LINTUL5NPKDemand, SlimNitrogen, LINTUL5Biomass, SlimRoots, and SlimWater, amongst others. 
An overview of key SimComponents\footnote{More information about SIMPLACE components: \url{https://www.simplace.net/index.php/documentation}} is described in Seidel et al. \cite{seidel2019estimation}. 
Specifically, the biomass per species was calculated by SimComponent LINTUL5Biomass, which considers water and nitrogen limitation effects on biomass increment. 
The mixture model was developed in the SIMPLACE framework and simulates the splitting of solar radiation according to the competition of the two species as well as the water and nitrogen uptake of two crop species that are planted in a mixture. 
The model was calibrated and tested on three environments (CKA 2020, 2021, and WG 2020) based on collected data from the crops cultivated solely and evaluated based on the data in the mixture treatments.

\section{Results and Discussions}\label{sec:results}
In this section, the results of the growth estimation models are described at the beginning, as the accuracies of these models are needed for the discussion of the image prediction results.
In the following, we first show the results of image prediction with only temporal variation, which allows a comparison with reference data, then simulations with further changed conditions, and finally, the transferability to another experimental site.

\subsection{Accuracy of projected leaf area estimation}\label{sec:acc_pla}

\begin{table*}[t]
\centering
\begin{tabular}{lrrrrrrrr}
\hline
            & \multicolumn{4}{l}{Bounding Box} & \multicolumn{4}{l}{Segmentation} \\
            & AP    & AP@0.50 & AP@0.75 & AR   & AP    & AP@0.50 & AP@0.75 & AR   \\
\hline
Arabidopsis & 0.92  & 0.99    & 0.99    & 0.95 & 0.77  & 0.99    & 0.98    & 0.78 \\
GrowliFlower & 0.86  & 0.96    & 0.92    & 0.88 & 0.78  & 0.97    & 0.92    & 0.82 \\
\hline
\end{tabular}
\caption{Mask R-CNN instance segmentation accuracies for the real (non-generated) images of the test set divided into bounding box and segmentation. Overall average precision (AP), with thresholds at $\text{IoU}=0.50$ and $\text{IoU}=0.75$, and overall average recall (AR) are given.}
\label{tab:segm_accuracy}
\end{table*}
Instance segmentation, which is used to derive PLA (projected leaf area), is trained on a small subset of the corresponding datasets for which reference segmentation masks are available.
Exact numbers for all datasets can be found in the bottom part of \tabref{tab:data_stats}.
The reference masks of the test set specified there are used to run the evaluation in \tabref{tab:segm_accuracy}.
It shows the instance segmentation accuracies using the measures AP and AR, which - due to the direct derivation from these - correlates with the accuracy of the PLA.
The GrowliFlower accuracies are comparable to the results of Kierdorf et al. \cite{kierdorf2023growliflower}, i.e., sufficient to evaluate cauliflower growth.
Arabidopsis has a higher AP and AR for bounding boxes and is at a comparable high level to GrowliFlower for segmentation, thus also adequate to determine PLA.

\subsection{Accuracy of biomass estimation}\label{sec:acc_biomass}

\begin{table*}[t]
\centering
\begin{tabular}{cccrrrrrrrr}
\hline
& & \multicolumn{2}{c}{SW/FB Mix} & \multicolumn{2}{c}{SW mono} & \multicolumn{2}{c}{FB mono} & \multicolumn{2}{c}{Overall} \\
& & MAE & ME & MAE & ME & MAE & ME & MAE & ME \\
\hline
\multirow{2}{*}{Mixed-CKA} & SW  & 0.142 & -0.006 & 0.188 & -0.074 & 0.001 &  0.001 & 0.142 & -0.026 \\
                           & FB  & 0.125 & -0.008 & 0.017 & 0.017  & 0.179 &  0.052 & 0.097 & 0.005  \\
\multirow{2}{*}{Mixed-WG}  & SW  & 0.126 & -0.023 & 0.150 & -0.050 & 0.018 &  0.018 & 0.122 & -0.027 \\
                           & FB  & 0.105 & -0.026 & 0.003 & 0.003  & 0.185 & -0.046 & 0.082 & -0.019 \\
\hline
\end{tabular}
\caption{Biomass estimation accuracies assessed by MAE and ME between the predictions from the real (non-generated) images of the test set and reference values from the process-based crop growth model. The scores are separated for mixtures and SW resp. FB monocultural fields. All units are given in \SI{}{t/ha}.}
\label{tab:bm_regress_results}
\end{table*}

\begin{figure*}[t]
    \centering
    \includegraphics[width=1.0\textwidth]{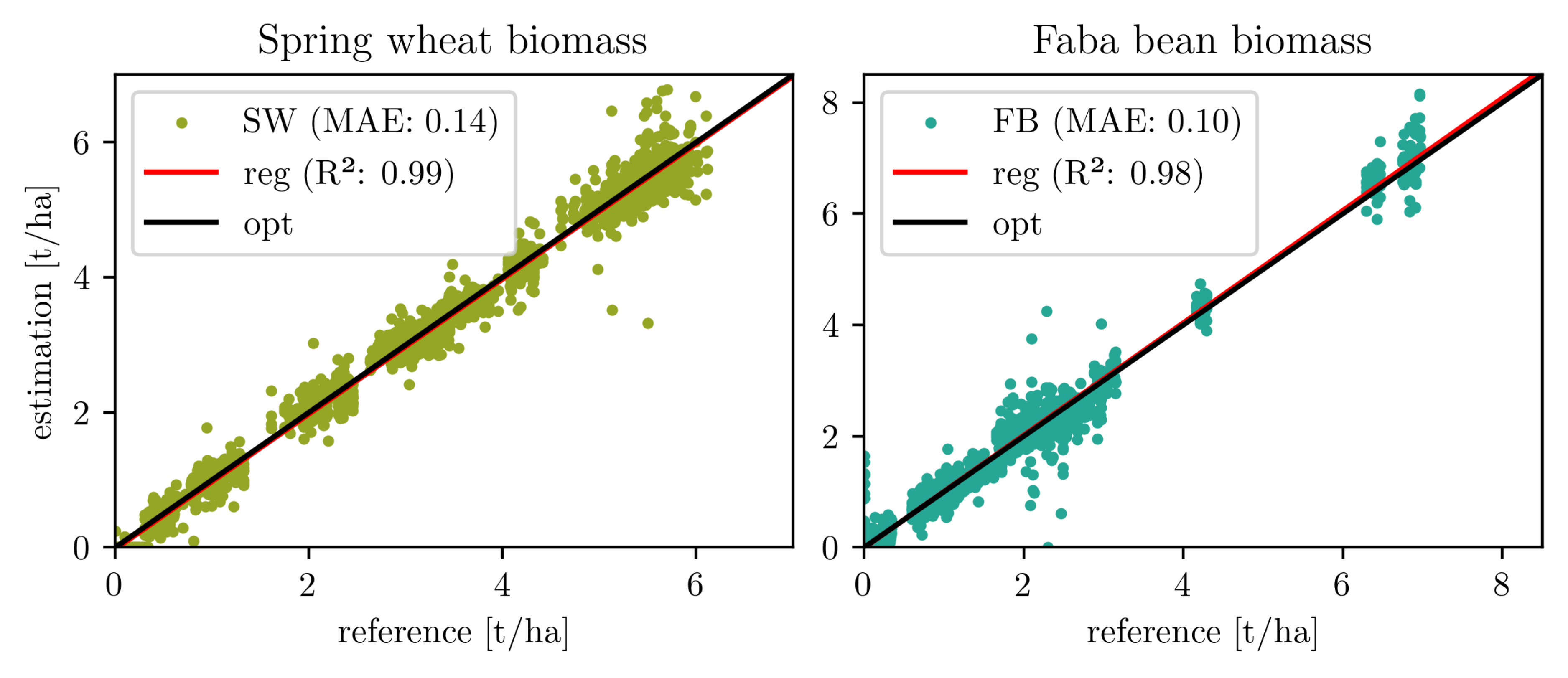}
    \caption{Scatter results of dried biomass estimation from Mixed-CKA imagery over all growth stages and all treatments (mixtures and monocultural fields) split up in spring wheat (SW) and faba bean (FB) for Mixed-CKA.}
    \label{fig:mix_reg_result}
\end{figure*}

The accuracy of dried biomass estimation for both MixedCrop sites is given in \tabref{tab:bm_regress_results}.
For mixtures the MAE is between \SI{0.126}{t/ha} and \SI{0.142}{t/ha} for SW and between \SI{0.105}{t/ha} and \SI{0.125}{t/ha} for FB.
Notably, the ME is less than \SI{-0.01}{t/ha} for mixtures at CKA and less than \SI{-0.03}{t/ha} at WG for both species.
For the monoculture reference fields, the MAE is \SI{0.179}{t/ha} for FB in the FB monocultures and \SI{0.188}{t/ha} for SW in the SW monocultures.
This is slightly higher than in the mixtures, which is expected because in the monocultures, more of each species grows in absolute terms than in the mixtures. 
In return, the mixtures generally have a higher total biomass \cite{paul2023effects}.
The low estimation of SW on FB monocultures between \SI{0.001}{t/ha} and \SI{0.018}{t/ha} and vice versa FB on SW monocultures between \SI{0.003}{t/ha} and \SI{0.017}{t/ha} can be considered as additional evidence that the model is able to distinguish the species with high accuracy. 
It can be assumed that a common weed found in both fields, \textit{Chenopodium album}, which bears partial similarity to FB, is often incorrectly identified as FB. The mean absolute error (MAE) will be lower if there are fewer weeds or if it is included in the growth estimation model.

In \figref{fig:mix_reg_result}, the overall results for CKA are visualized as two scatter plots for SW and FB, where the estimations are plotted against the reference from the process-based crop growth model.
The regression line is close to the optimal line with a minimal underestimation for SW ($\text{ME}=\SI{-0.026}{t/ha}$) and a minimal overestimation ($\text{ME}=\SI{0.005}{t/ha}$) for FB.
In total, the regression results are $\text{MAE}=\SI{0.14}{t/ha}$ and ${\mathit{R}}^{2}=0.99$ for SW and $\text{MAE}=\SI{0.10}{t/ha}$ and ${\mathit{R}}^{2}=0.98$ for FB.
With this, the model is considered as accurate enough for an evaluation of generated images.

When assessing the following results, it is important to consider that the accuracy of the predicted images heavily relies on the accuracy of the growth estimation models. 
The accuracy of these models is evaluated solely based on real reference images. 
Any discrepancy between the growth estimation of these real reference images and the predicted images can be attributed to two factors. 
Firstly, it could be due to actual differences in plant phenotypes compared to the reference images. This is the deviation we aim to identify. 
Secondly, part of the deviation may be caused by potential small corruptions or artifacts in the artificial images, even if they pass GAN evaluation metrics. 
These corruptions can lead to incorrect predictions by the growth estimation model despite the visible plant phenotypes in the artificial images being accurate. This is because the growth estimation model was not trained on corrupted images. 
While it is not possible to completely avoid or quantify the second source of deviation, we strive to minimize it by augmenting the data used to train the growth estimation model, making it more robust and less susceptible to corruption.

\subsection{Time-varying image prediction}\label{sec:results_img_t}

\begin{table*}[t]
\centering
\begin{tabular}{lcrrrrrr}
\toprule
\multirow{2}{*}{} & Train conds. & \multicolumn{4}{c}{MS-SSIM (\textuparrow)} & \multicolumn{1}{c}{LPIPS (\textdownarrow)} & \multicolumn{1}{c}{FID (\textdownarrow)}\\
& t\hspace{2mm}trt\hspace{2mm}bm & \multicolumn{1}{c}{$\text{T}_0$} & \multicolumn{1}{c}{ST} & \multicolumn{1}{c}{LT} & \multicolumn{1}{c}{\o} & \multicolumn{1}{c}{\o} & \multicolumn{1}{c}{\o}\\
\midrule
Arabidopsis & \checkmark\hspace{3mm}\texttimes\hspace{3mm}\texttimes & 0.94 & 0.81 & 0.68 & 0.80 & 0.25 & 6.54\\
GrowliFlower & \checkmark\hspace{3mm}\texttimes\hspace{3mm}\texttimes & 0.98 & 0.30 & 0.20 & 0.29 & 0.51 & 20.17\\
Mixed-CKA & \checkmark\hspace{3mm}\texttimes\hspace{3mm}\texttimes & 0.99 & 0.23 & 0.22 & 0.30 & 0.46 & 20.44\\
Mixed-CKA & \checkmark\hspace{3mm}\checkmark\hspace{3mm}\texttimes & 0.97 & 0.25 & 0.23 & 0.31 & 0.47 & 16.26\\
Mixed-CKA & \checkmark\hspace{3mm}\checkmark\hspace{3mm}\checkmark & 0.99 & 0.23 & 0.22 & 0.29 & 0.46 & 24.86\\
\midrule
Mixed-WG\footnotemark[1] & \checkmark\hspace{3mm}\texttimes\hspace{3mm}\texttimes & 0.92 & 0.13 & 0.11 & 0.20 & 0.50 & 40.67\\
\bottomrule
\end{tabular}
\footnotetext[1]{Transferability check: Model trained on Mixed-CKA and applied to Mixed-WG.}
\caption{Evaluation with metrics MS-SSIM, LPIPS, and FID. Each row represents a distinct model, each trained with an input image along with the specified conditions; for testing, only the input image and t are varied. MS-SSIM is reported for generations with different $\lvert\Delta t\rvert$ filters: $\text{T}_0$: identity $\lvert\Delta t\rvert=0$; ST: short-term $1\leq\lvert\Delta t\rvert\leq 10$; LT: long-term $\lvert\Delta t\rvert\geq11$.}
\label{tab:gen_scores_classic}
\end{table*}

\begin{table*}[t]
\centering
\begin{tabular}{lcrrrrr}
\toprule
\multirow{2}{*}{} & Train conds. & \multicolumn{4}{c}{MAE} & \multicolumn{1}{c}{ME} \\
& t\hspace{2mm}trt\hspace{2mm}bm & \multicolumn{1}{c}{$\text{T}_0$} & \multicolumn{1}{c}{ST} & \multicolumn{1}{c}{LT} & \multicolumn{1}{c}{\o} & \multicolumn{1}{c}{\o} \\
\midrule
Arabidopsis & \checkmark\hspace{3mm}\texttimes\hspace{3mm}\texttimes & 0.27 & 0.76 &  1.44 & 0.82 & -0.32 \\
GrowliFlower & \checkmark\hspace{3mm}\texttimes\hspace{3mm}\texttimes & 6.41 & 8.84 & 10.18 & 9.64 & 1.27  \\
\bottomrule
\end{tabular}
\caption{Plant-specific evaluation with projected leaf area (PLA) assessed by MAE and ME in the unit \%/image. MAE is reported for generations with different $\lvert\Delta t\rvert$ filters: $\text{T}_0$: identity $\lvert\Delta t\rvert=0$; ST: short-term $1\leq\lvert\Delta t\rvert\leq 10$; LT: long-term $\lvert\Delta t\rvert\geq11$.}
\label{tab:gen_scores_pla}
\end{table*}

\begin{table*}[t]
\centering
\begin{tabular}{lccrrrrrrr}
\toprule
\multirow{3}{*}{} & Train & \multirow{3}{*}{} & \multicolumn{4}{c}{OA} & \multicolumn{1}{c}{OA} & \multicolumn{1}{c}{Mix} & \multicolumn{1}{c}{Mix} \\
& conds. &  & \multicolumn{4}{c}{MAE} & \multicolumn{1}{c}{ME} & \multicolumn{1}{c}{MAE} & \multicolumn{1}{c}{ME}  \\
& t\hspace{2mm}trt\hspace{2mm}bm & & \multicolumn{1}{c}{$\text{T}_0$} & \multicolumn{1}{c}{ST} & \multicolumn{1}{c}{LT} & \multicolumn{1}{c}{\o} & \multicolumn{1}{c}{\o}       & \multicolumn{1}{c}{\o}   & \multicolumn{1}{c}{\o} \\
\midrule
\multirow{2}{*}{Mixed-CKA} & \multirow{2}{*}{\checkmark\hspace{3mm}\texttimes\hspace{3mm}\texttimes} & SW & 0.22 & 0.42 & 0.39 & 0.38 &  0.12 & 0.31 &  0.20 \\
                           &                                                                         & FB & 0.16 & 0.34 & 0.30 & 0.28 & -0.12 & 0.25 & -0.17 \\
\multirow{2}{*}{Mixed-CKA} & \multirow{2}{*}{\checkmark\hspace{3mm}\checkmark\hspace{3mm}\texttimes} & SW & 0.30 & 0.22 & 0.25 & 0.24 &  0.09 & 0.25 & 0.15 \\
                           &                                                                         & FB & 0.24 & 0.16 & 0.19 & 0.19 & -0.13 & 0.24 & -0.15 \\
\multirow{2}{*}{Mixed-CKA} & \multirow{2}{*}{\checkmark\hspace{3mm}\checkmark\hspace{3mm}\checkmark} & SW & 0.17 & 0.21 & 0.18 & 0.18 & -0.02 & 0.18 &  0.05 \\
                           &                                                                         & FB & 0.11 & 0.16 & 0.14 & 0.13 & -0.01 & 0.15 & -0.04 \\
\midrule
\multirow{2}{*}{Mixed-WG}\footnotemark[1]  & \multirow{2}{*}{\checkmark\hspace{3mm}\texttimes\hspace{3mm}\texttimes} & SW & 0.45 & 1.25 & 1.14 & 1.07 & 0.18 & 1.06 & 0.24 \\
                           &                                                                                         & FB & 0.41 & 0.48 & 0.67 & 0.64 & -0.04 & 0.62 & -0.11 \\
\bottomrule
\end{tabular}
\footnotetext[1]{Transferability check: Model trained on Mixed-CKA and applied to Mixed-WG.}
\caption{Plant-specific evaluation with biomass (BM) assessed by MAE and ME in the unit t/ha given for all (OA) and mixture (Mix) fields and divided into spring wheat (SW) and faba bean (FB) biomasses. Overall MAE is reported for generations with different $\lvert\Delta t\rvert$ filters: $\text{T}_0$: identity $\lvert\Delta t\rvert=0$; ST: short-term $1\leq\lvert\Delta t\rvert\leq 10$; LT: long-term $\lvert\Delta t\rvert\geq11$.}
\label{tab:gen_scores_bm}
\end{table*}

The first image prediction experiment will evaluate how accurately our framework predicts images of other growth stages of the plant, given an input image and a different amount of conditions used for training, as indicated in \tabref{tab:gen_scores_classic}. 
For each prediction, conditions are used that match the input image and a randomly picked time-varying reference image of this plant.
Multiple models are trained on the different datasets and with a varying combination of conditions, namely time (t), treatment (trt), and simulated biomass (bm).

In \tabref{tab:gen_scores_classic}, the predicted image quality is evaluated using the metrics MS-SSIM, LPIPS, and FID. 
Across all predictions, the highest accuracies are obtained with the Arabidopsis dataset for all three metrics MS-SSIM=0.8, LPIPS=0.25, and FID=6.54, while similarly lower overall accuracies are obtained with the GrowliFlower and MixedCrop datasets. 
For these, the MS-SSIM is between 0.29 and 0.31, LPIPS is between 0.46 and 0.51, and FID is between 16.26 and 24.86.
Particularly remarkable is the dependence of the accuracy on the prediction distance, where MS-SSIM is higher for all datasets, the smaller $\lvert\Delta t\rvert$.
In the case of $\Delta t = 0$, the model acts as an autoencoder, meaning that it reproduces the input, also known as identity mapping. 
In this case, the results show an MS-SSIM of 0.94 for Arabidopsis and MS-SSIM values between 0.97 and 0.99 for the Mixed-CKA models.
From short-term (ST) to long-term (LT) predictions, the MS-SSIM decreases continuously up to 0.20.

It is noticeable that Arabidopsis has better values in all metrics except $\text{T}_0$ than GrowliFlower and Mixed-CKA, which can be attributed to the daily recording times and controlled laboratory conditions with constant light and no weather effects.
The identity mapping ($\text{T}_0$) is worse than the other datasets because in the Arabidopsis dataset, multiple images were taken per day, which means it is not a strict identity mapping. 
However, this can be altered by changing the model time unit from days to hours.
The MS-SSIM decrease from $\text{T}_0$ over ST to LT means the less far the model predicts into the future or past, the better the predicted images match the reference.
Particularly, an MS-SSIM below 0.3 implies less similarity between predicted and reference images.
In parallel, the FID for all models, including ST and LT predictions, is below 25, which can be considered as good image quality.
This is expected because, with increasing prediction steps, detailed plant appearances, like leaf counts and orientations, are increasingly difficult to predict, while general structural traits, like plant positions and overall sizes, can be predicted more accurately.

Insight into the usability of predicted images can be drawn from the plant-specific evaluation results using projected leaf area (PLA) estimation for Arabidopsis and GrowliFlower and biomass (BM) estimation for MixedCrop.
\tabref{tab:gen_scores_pla} shows the obtained results for Arabidopsis and GrowliFlower in \tabref{tab:gen_scores_pla}. It can be seen that MAE increases with larger $\lvert\Delta t\rvert$ in both cases, but the overall accuracy of \SI{<1}{\%} is high for Arabidopsis and with \SI{<10}{\%} slightly lower for GrowliFlower.
In addition, for Arabidopsis, a mean error of $\SI{-0.32}{\%}\approx\SI{-11}{mm^2}$ indicates a small mean underestimation, while GrowliFlower heads are predicted larger $\text{ME}=\SI{1.27}{\%}\approx\SI{80}{cm^2}$ than the corresponding reference.
The biomass evaluation for Mixed-CKA in \tabref{tab:gen_scores_bm} is divided into models trained with different conditions.
All scores are given separately for SW and FB; moreover, an average over all field plots and all mixture plots is reported.
The MAE separation into different prediction distances shows that for $\text{T}_0$ the lowest deviations occur with a small increase to ST, but a decrease (accuracy gain) for LT over ST.
The overall MAE ranges from \SI{0.13}{t/ha} to \SI{0.38}{t/ha} and is comparable to Mix MAE, where only the crop mixture field plots are integrated.
Thereby, overall SW MAE is always higher than FB MAE with a magnitude of up to \SI{0.1}{t/h}.
Noticeably, overall FB ME is negative while SW ME is positive for all models except those trained on all conditions, showing a systematic SW over- and a FB underestimation. 
With an increasing number of conditions, the overall MAE decreases significantly by \SI{0.2}{t/ha} for SW and \SI{0.15}{t/ha} for FB.
Comparing the accuracy when biomass estimation is performed on predicted mixtures (last two columns of \tabref{tab:gen_scores_bm}) with the accuracy when it is performed on real mixtures (first two columns of \tabref{tab:bm_regress_results}) two results are shown:
First, the MAE of the predicted mixtures using the model with all conditions is slightly above the MAE of the real mixtures (SW: +0.04 t/ha, FB: +0.03 t/ha).
The other models trained with fewer conditions show higher deviations up to \SI{+0.17}{t/ha} for SW and \SI{+0.13}{t/ha} for FB.  
Second, the ME of the predicted mixtures using the model with all conditions is by a magnitude of 5 above the ME of the real mixtures.

We provide two assumptions for the SW and FB differences in MAE and ME:
We assume that having for SW a generally higher MAE magnitude than for FB is caused by the higher absolute SW biomass level in the field.
Additionally, we assume the reason for the systematic overestimation of SW and underestimation of FB (indicated by ME) is due to the unbalanced dataset: there are significantly more SW than FB monocultures. 
We argue that the image prediction model copes worse with this unbalanced dataset than the growth estimation model.
Besides, MAE and ME decrease significantly as more conditions are added to the model.
This can be explained by the model being better informed about the crop growth behavior if it receives more growth-influencing factors and can thus become more accurate.
There is a loss of accuracy from identity mapping to short-term predictions but no significant loss of accuracy from short-term to long-term predictions.
Thus, long-term predictions can be considered valuable for phenotyping applications.

So, the quantitative evaluation leads to the overall finding: 
Although the predicted images do match the reference images less at large $\lvert\Delta t\rvert$, they represent realistic plants of their respective growth stage, as indicated by FID, and are still accurate enough to derive reasonable plant traits, as indicated by plant-specific evaluation.

\begin{figure*}[t]
    \centering
    \includegraphics[width=1.0\textwidth]{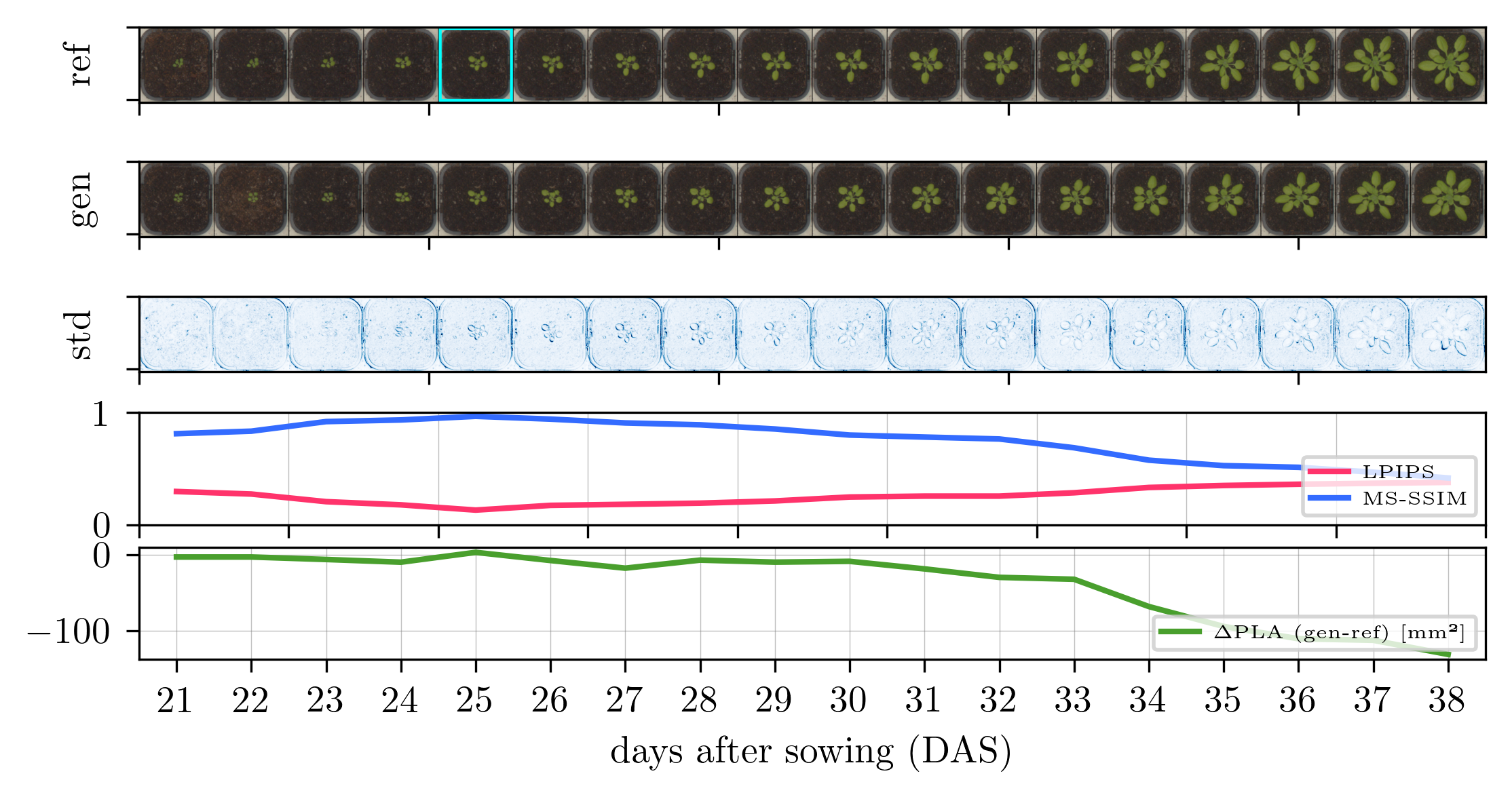}
    \caption{Time-varying image prediction for Arabidopsis with, in the top row, reference images with an early growth stage as input (cyan frame), in the second row, all day-wise generated predictions, and, in the third row, standard deviation images over 10 predictions with different noise input \d{z} and otherwise constant input conditions. The two bottom rows have the quality metrics: learned perceptual image patch similarity (LPIPS), multiscale structural similarity (MS-SSIM), and the projected leaf area difference ($\Delta$PLA)}
    \label{fig:qual_abd_img_t}
\end{figure*}

\begin{figure*}[t]
    \centering
    \includegraphics[width=1.0\textwidth]{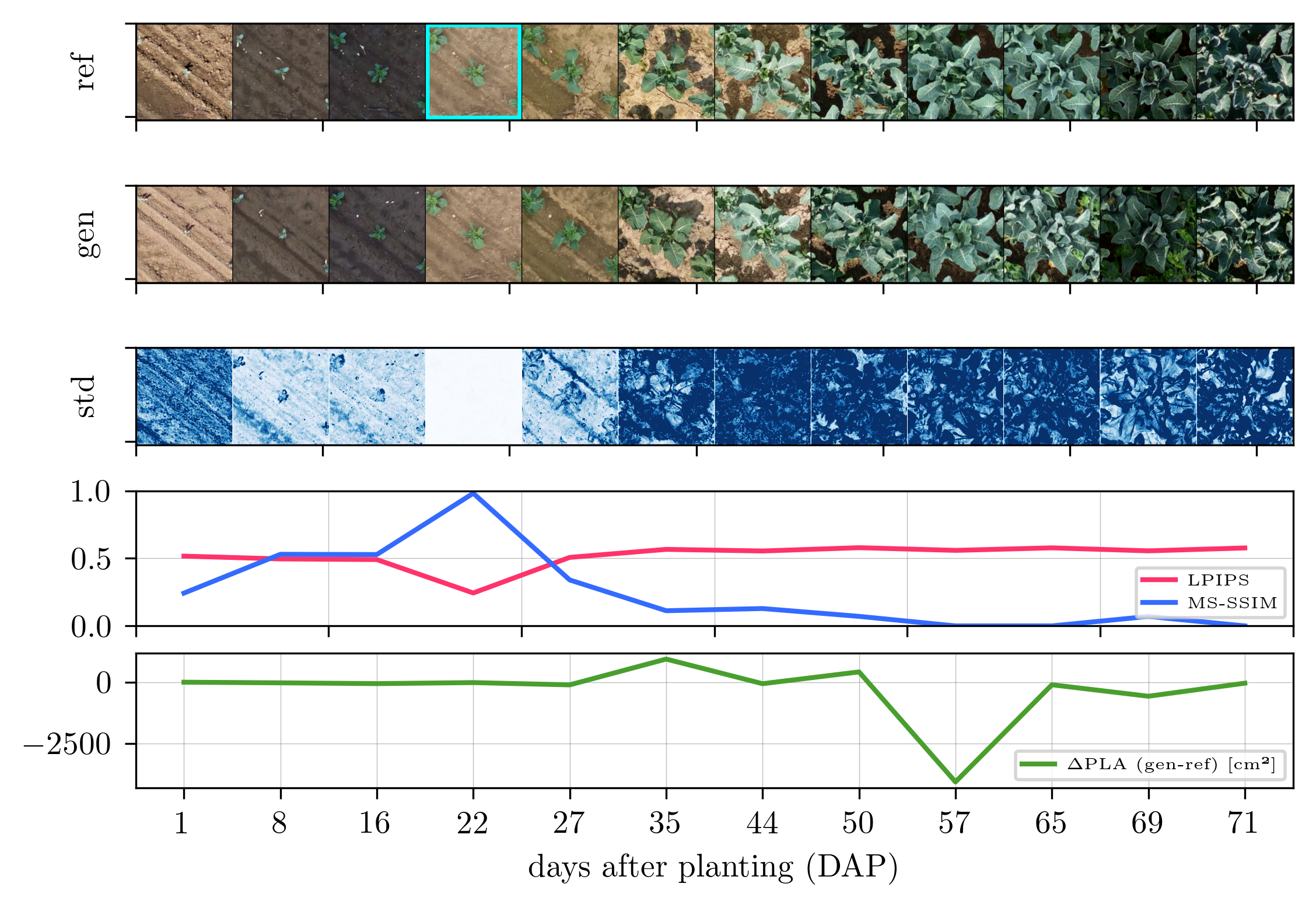}
    \caption{Time-varying image prediction for GrowliFlower with, in the top row, reference images with an early growth stage as input (cyan frame), in the second row, all day-wise generated predictions, and, in the third row, standard deviation images over 10 predictions with different noise input \d{z} and otherwise constant input conditions. The two bottom rows show the quality metrics: learned perceptual image patch similarity (LPIPS), multiscale structural similarity (MS-SSIM), and the projected leaf area difference ($\Delta$PLA)}
    \label{fig:qual_grf_img_t}
\end{figure*}

\begin{figure*}[t]
    \centering
    \includegraphics[width=1.0\textwidth]{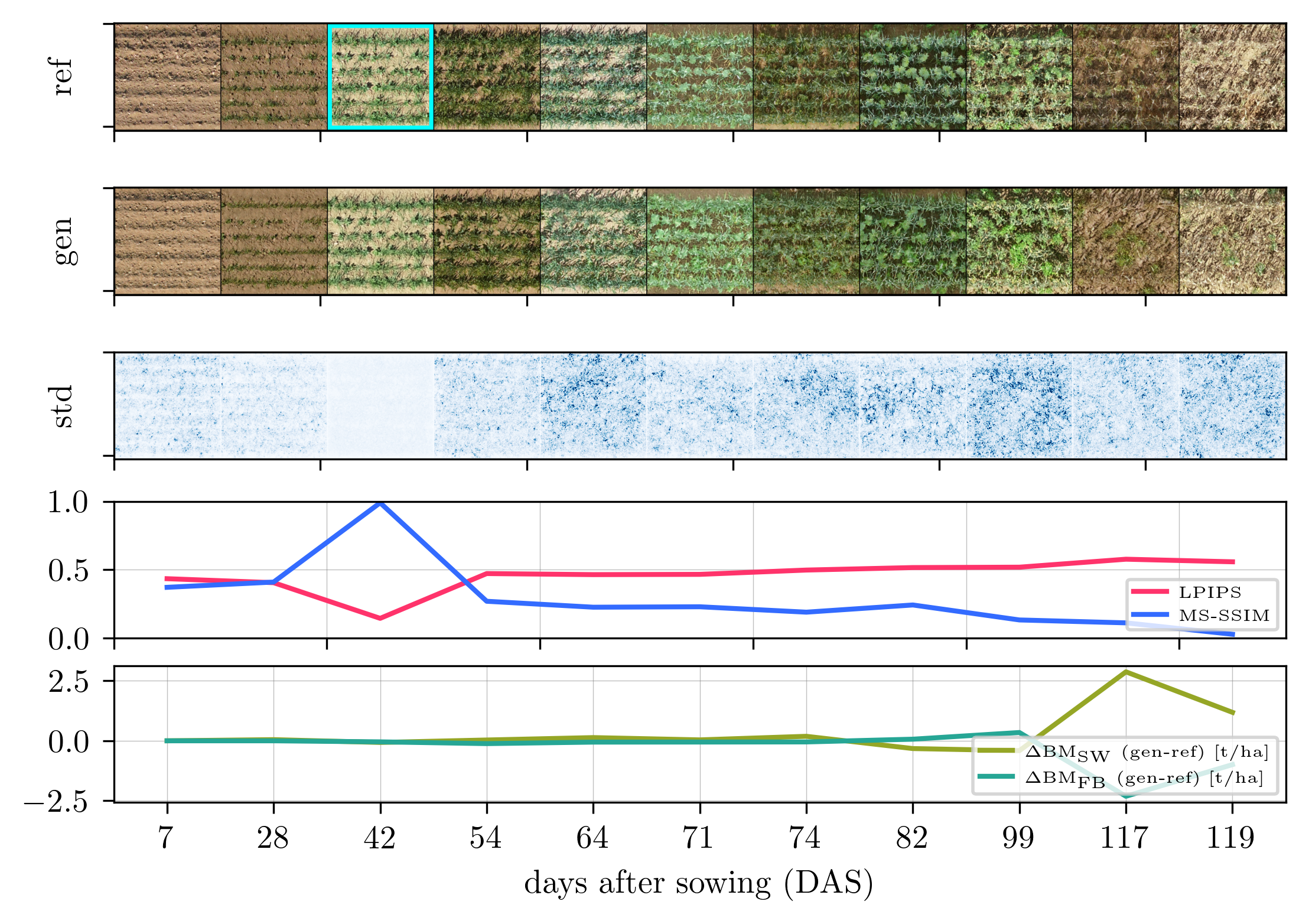}
    \caption{Time-varying image prediction for Mixed-CKA with, in the top row, reference images with an early growth stage as input (cyan frame), in the second row, all day-wise generated predictions, and, in the third row, standard deviation images over 10 predictions with different noise input \d{z} and otherwise constant input conditions. The two bottom rows show the quality metrics: learned perceptual image patch similarity (LPIPS), multiscale structural similarity (MS-SSIM), and the biomass differences for spring wheat ($\Delta \text{BM}_\text{SW}$) and faba bean ($\Delta \text{BM}_\text{FB}$)}
    \label{fig:qual_mix_img_t}
\end{figure*}

Further findings can be drawn from qualitative results showing hand-picked time-varying image prediction results in \figref{fig:qual_abd_img_t} for Arabidopsis, \figref{fig:qual_grf_img_t} for GrowliFlower, and \figref{fig:qual_mix_img_t} for Mixed-CKA, where models are used that are trained on the temporal condition only.
Each figure consists of 5 rows: 
The first row contains a reference plant over time, where an early growth stage with a cyan frame is the input to the model in each case.
The second row shows generated images by keeping except time all other conditions, including noise \d{z}, constant.
The third row shows the variability image, which is the standard deviation over 10 predictions of the same time point with different \d{z}, whereby the standard deviation is averaged over all RGB channels and overdrawn by a factor of four for clearer visualization.
The darker the blue, the greater the variability for each pixel within the 10 predictions.
The fourth and fifth rows show the classical and plant-specific evaluation metrics for each gen-ref image pair.

For all datasets and time points, the predictions are realistic, with a few exceptions, such as the last image of GrowliFlower.
Comparing the variability images, Arabidopsis has the lowest pixel-wise standard deviation, followed by MixedCrop and GrowliFlower.
In all cases, there is high variability at the leaf edges, where the actual uncertainty is greatest.
The LPIPS and MS-SSIM deteriorate with increasing $\Delta t$ with a peak each for identity mapping. 
Plant property curves differ for each data set:
In Arabidopsis, $\Delta$PLA is close to zero until \SI{30}{DAS} and then drifts into the negative range, indicating a leaf area underestimation for advanced growth stages.
In GrowliFlower, the curve is close to zero with small fluctuations except for a large negative peak at 57 DAP, indicating that the leaf area could not be correctly estimated from the predicted image of this day.
Similarly, for Mixed-CKA, the curves stay around zero until day 99, after which SW biomass is significantly overestimated with up to \SI{+2.5}{t/ha} and FB biomass is significantly underestimated with up to \SI{-2.5}{t/ha}.

There are two important insights that emerge from the visualized images.
First, a strong consistency of the generated images over time is given, which is visible in Arabidopsis and GrowliFlower through leaf orientations but also through neighboring plants and in Mixed-CKA through certain crop patterns such as small gaps (second crop row, right) or weeds (third and fourth crop row, center).
Second, the dependence of the generated images on the input is visible for all datasets, particularly in the position of the plants and crop rows and by granules on the ground, which can be found on the input image as well as on several generated images.
While the variability images show realistic uncertainties at the leaf edges, they also reveal a limitation in the image prediction:
While the identity mapping has no or extremely low variability, as expected, no continuous increase in variability over time is evident, leading to overconfidence at large $\Delta t$ where variability would be expected to be significantly higher.
The parallel examination of MS-SSIM and LPIPS with the images confirms the findings from the quantitative results: 
Despite the images being less consistent with the reference as the prediction distance increases, there is neither a general visual quality decrease nor a general decrease in the accuracy of the estimated plant traits for time-varying predictions.

An overview of predictions for days not present in the datasets, so temporally out-of-distribution (OOD) can be found in Appendix \ref{secA:t-ood}.
While challenging due to large spectral differences between images of existing time points, it can be shown that realistic images can still be generated at new time points.

\subsection{Data-driven simulation using treatment information}
\begin{figure*}[t]
    \centering
    \includegraphics[width=0.95\textwidth]{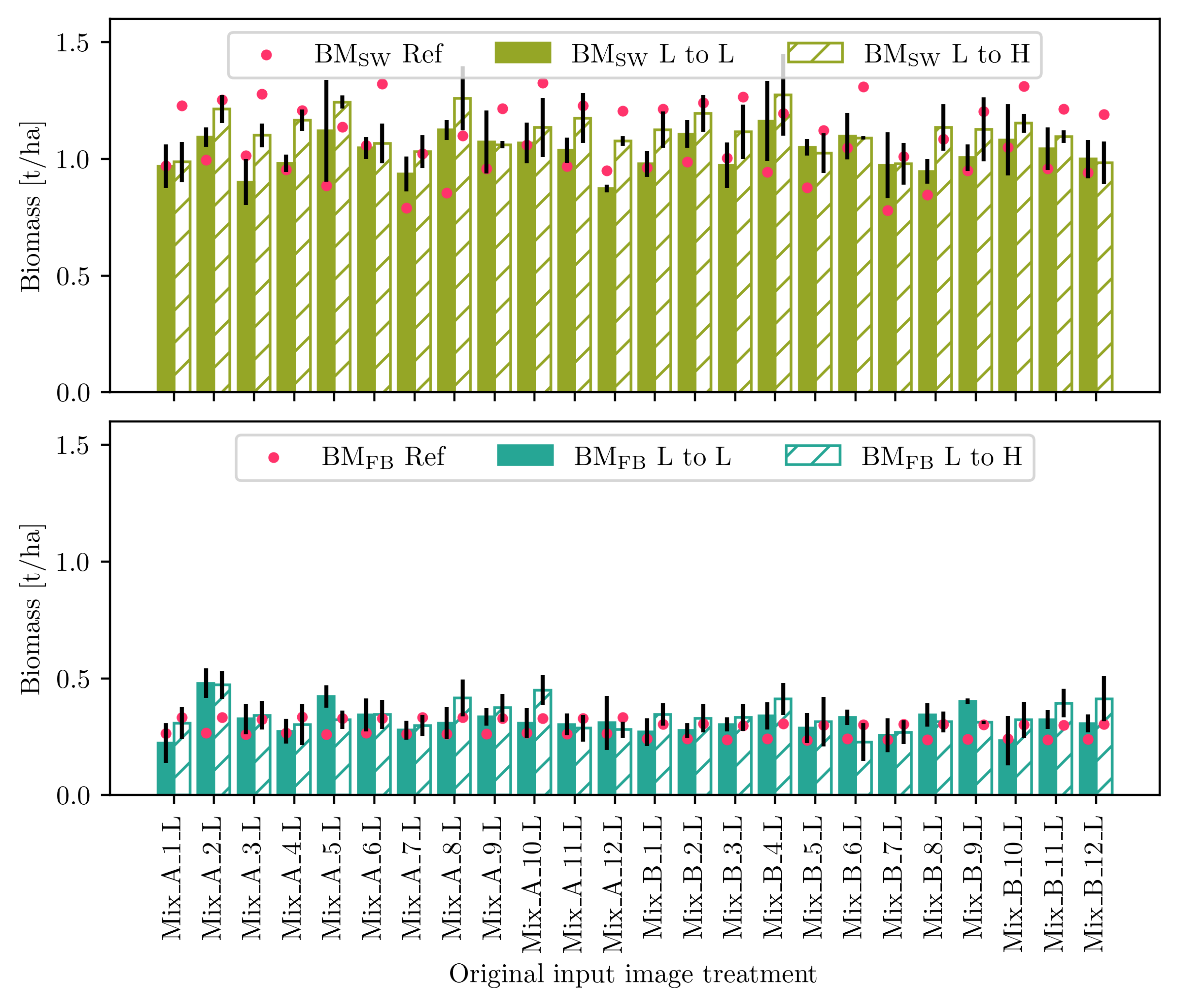}
    \caption{Simulating the SW (top) and FB (bottom) change from a low (L) density to a high (H) density treatment for all mixture field plots and the growth prediction step 28 DAS to 54 DAS. While filled bars represent the comparative prediction under the original treatment, hashed bars represent the simulated treatment change. Black lines symbolize the standard deviation across treatment replicates; red dots symbolize the outcome of the process-based crop growth model for the resp. treatments and 54 DAS.}
    \label{fig:simulation_LH}
\end{figure*}

\begin{figure*}[t]
    \centering
    \includegraphics[width=0.95\textwidth]{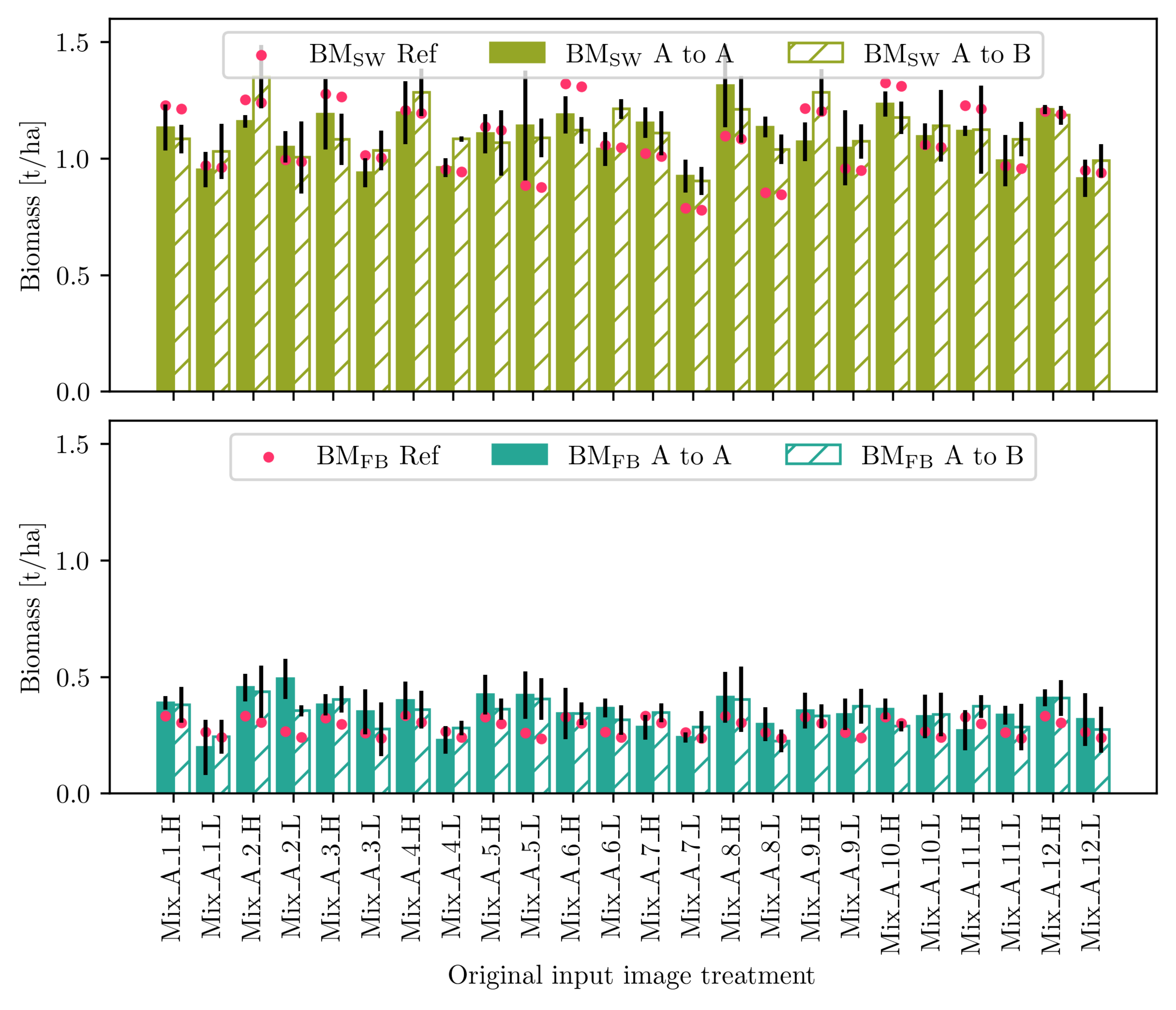}
    \caption{Simulating the SW (top) and FB (bottom) change from faba bean cultivar \textit{Mallory} (A) to cultivar \textit{Fanfare} (B) for all mixture field plots and the growth prediction step 28 DAS to 54 DAS. While filled bars represent the comparative prediction under the original treatment, hashed bars represent the simulated treatment change. Black lines symbolize the standard deviation across treatment replicates; red dots symbolize the outcome of the process-based crop growth model for the resp. treatments and 54 DAS.}
    \label{fig:simulation_AB}
\end{figure*}

The data-driven simulations on the MixedCrop dataset are intended to show the flexibility of the image-prediction model in the presence of changing growth-influencing variables.
To enable an illustrative and informative demonstration and visualization, we systematically vary the time (t) and treatment (trt) information as a condition for the Mixed-CKA dataset. 
We use the results to investigate and evaluate how different treatments appear in the future when something about the treatment changes starting from a certain initial condition (image). 
We would like to emphasize that the performed change in treatments is intended for the evaluation of the method and is thus limited in its realistic nature, yet aims to show that our framework is applicable to realistic scenarios. 
Our expectation is that the estimated biomass from the data-driven simulation changes in the same direction as that of the process-based plant growth model, confirming the reliable image predictions of our system.

In particular, two simulations are conducted from the input time point of \SI{28}{DAS} to \SI{54}{DAS} where first, the seed density is changed from low (L) to high (H) (\figref{fig:simulation_LH}), and second, the faba bean cultivar is changed from Mallory (A) to Fanfare (B) (\figref{fig:simulation_AB}).
Thus, the input image is encoded in the original treatment, but a treatment change is made to decode the simulated future plant phenotype.
The figures compare the data-driven prediction without treatment change (filled bars) with the prediction including treatment change (hashed bars) and the process-based predictions for the respective target treatment (red dots).
Since there are multiple replicates for each treatment, the bars represent the mean, and the black lines represent the standard deviation.
We deliberately chose an early stage as the input because the differences in biomass between the treatments are not yet too great, and differences between the FB varieties are hardly discernible.
However, we do not use DAS=7, which is bare soil, because we do want to observe the spatial development of the crops.
In addition, we focus on mixtures in the simulations to be able to analyze the biomass of spring wheat and faba bean in parallel.

Focusing on the simulation of L$\rightarrow$H in \figref{fig:simulation_LH}, the data-driven estimated biomass of the high-density simulated treatments (hashed bars) is higher than that of the low-density simulated ones (filled bars) for SW in 20/24 cases and for FB in 16/24 cases.
The process-based biomass gain from L$\rightarrow$H, shown by the red dots, is for SW significantly higher (\SI{0.25}{t/ha}) than for FB (\SI{<0.1}{t/ha}).
On average, over all treatments, the biomass increases for both SW and FB. 
It is noticeable that FB biomass is slightly overestimated compared to the reference in almost all cases and that SW biomass is often overestimated for the L$\rightarrow$L simulation while underestimated for L$\rightarrow$H.

The analysis of the simulation of faba bean cultivar A$\rightarrow$B in \figref{fig:simulation_AB} is more challenging because only a small loss of biomass is expected for FB and an even smaller one for SW (almost the same level), as shown by the red dots.
Treatment-wise, this decrease is not clearly visible for either SW or FB. 
Only in slightly more than half of the treatments is the hashed bar smaller than the filled bar for both SW (13/24) and FB (15/24).
In average over all treatments, the hashed bars are smaller than the filled bares, albeit in the range of the standard deviation.
Comparing high and low-density treatments, it can be seen that the estimated biomass from the high-density treatments is higher for SW in 10/12 cases and for FB in 7/12 cases.

Both simulation results show that even small changes in the growth-influencing factors affect the predicted images.
Thereby, the reliability of the simulations is supported by the overall biomass increase from L$\rightarrow$H treatments and decrease from faba bean cultivar A$\rightarrow$B.
If the prediction change (filled to hashed bar) for individual treatments does not correspond to the expected change (red dots), there are three possible interpretations.
First, although the treatment condition is taken into account in the image prediction model, its influence might not be strong enough, so the differences in the generated images are not sufficiently prominent.
Second, the density resp. cultivar appearance of the input image might be already too prominent, making it difficult to change the growth stage at a later point; e.g., plants cannot arise from anywhere.
Third, the differences between low and high-density treatments resp. faba bean cultivars A and B are less clear in reality than the dynamic crop growth model suggests. 
In fact, the FB biomass gain for L$\rightarrow$H and the FB/SW biomass loss for A$\rightarrow$B is below the accuracy level of the biomass estimation (compare \tabref{tab:gen_scores_bm}, which can explain why a clear trend in biomass changes is not particularly apparent for these cases.
Another example where this simulation could be applied in practice is the prediction of weed pressure.
Apart from this experiment, we see the potential to simulate further treatment changes or their effects, e.g., weed cover.
This varies over the growing season and can be estimated quickly in categorical measures (low, medium, high), allowing crop growth predictions adapted to current field conditions.

\begin{figure*}[t]
    \centering
    \includegraphics[width=1.0\textwidth]{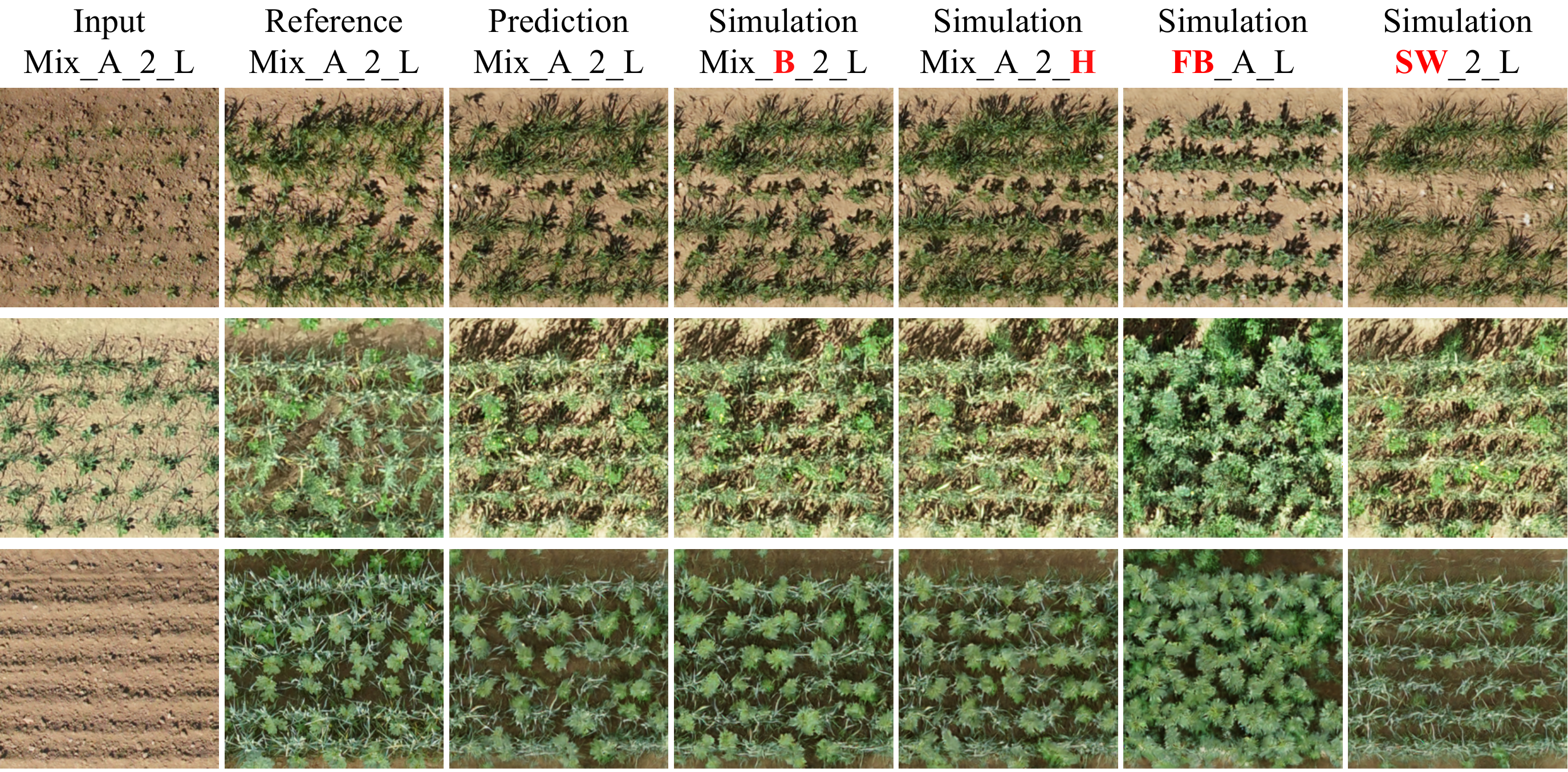}
    \caption{Growth simulation for different prediction steps and treatment changes in Mixed-CKA, first row \SI{28}{DAS} to \SI{45}{DAS}, second row \SI{42}{DAS} to \SI{99}{DAS}, and third row \SI{7}{DAS} to \SI{82}{DAS}. The first column shows the input image, the second the corresponding reference image of the future growth stage, the third the predicted image at these treatment conditions, and columns 4 to 7 show simulations of change in faba bean cultivar, density, and to monocultural reference.}
    \label{fig:simulation_trt_visual}
\end{figure*}
\figref{fig:simulation_trt_visual} also qualitatively illustrates the structural differences in the crop rows when simulating different treatments.
Besides the growth prediction step from \SI{28}{DAS} to \SI{54}{DAS}, two more growth prediction steps and two more treatment variations are simulated, including more unlikely scenarios, such as transformations of mixtures to monocultures.
While such simulations rarely make sense from an application point of view, as long as a mixture component is not completely suppressed, it is nevertheless noteworthy to see the model visualizing such a treatment change if necessary.

\subsection{Data-driven simulation using process-based biomass}
The following biomass simulation is intended to demonstrate the capability of including dynamic output variables of a process-based crop growth model in our framework.
For this, we use the trained Mixed-CKA model on time (t), treatment (trt), and process-based simulated biomass (bm), whereby the biomass systematically varied in order to get predictions for different possible SW and FB biomass ratios.
The time is randomly varied, so the simulation is performed over all growth stages by choosing a random prediction time point for each input mixture image and re-adjusting its biomass ratio.
The starting point for the simulation is the biomasses calculated dynamically from the process-based crop growth model for each time point and treatment, $\text{BM}_\text{SW}$ = $\text{BM}_\text{FB}$ = 100~\%.
While the image prediction model was trained with a fixed biomass value attached to each reference image, we will demonstrate that almost any combination of biomass ratios can be chosen for inference as long as they are within the range of the training data.

\begin{figure*}[t]
    \centering
    \includegraphics[width=1.0\textwidth]{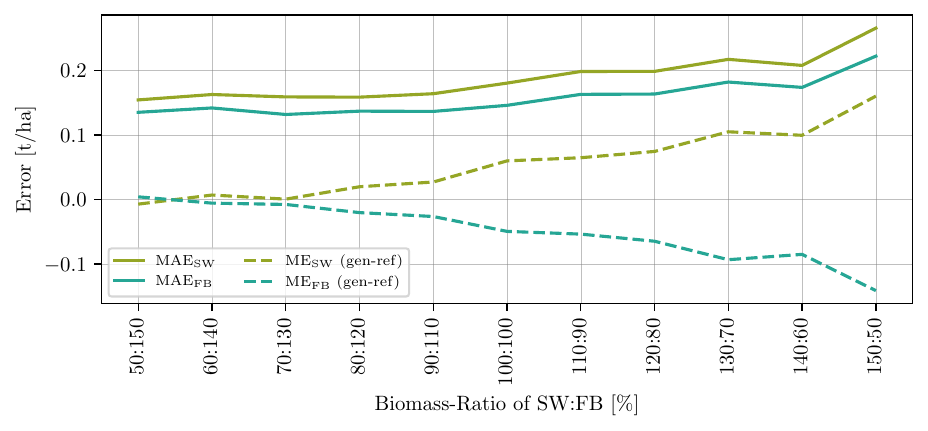}
    \caption{Comparing MAE and ME for image predictions from 28 DAS to 54 DAS with different expected spring wheat (SW) to faba bean (FB) biomass ratios.}
    \label{fig:simulation_BM}
\end{figure*}
\figref{fig:simulation_BM} shows MAE and ME respectively for SW and FB and different simulated biomass ratios, where the original composition (100:100) is shown in the middle, to the left, $\text{BM}_\text{FB}$ increases and to the right $\text{BM}_\text{SW}$.
This is accordingly also noticeable in the ME: 
If the BM fraction for SW and FB increases, more biomass is also estimated in the predicted image so that the ME increases.
So $\text{ME}_\text{SW}$ rises to the right, and the $\text{ME}_\text{FB}$ rises to the left.
The MAE reaches the minimum point at about the point where ME is also minimum at about the ratio 55~\%~SW to 145~\%~FB.

That a higher SW simulated biomass in the input of the framework also leads to a higher SW prediction in the output, for FB, accordingly, shows the reliability of our framework to generate predictions that realistically depend on the input conditions.
This demonstrates the capability of our framework to generate images that plausibly explain the output of a process-based model.
The reason why the minimum MAE/ME is not reached at the 100:100 state is mainly due to a slight dataset bias towards SW and the resulting under-prediction of FB plants in the images, as already discussed.
Assuming an unbiased image prediction model, we argue that this type of analysis can serve to improve the calibration of the process-based model and bring it closer to image-based field observations:
If the minimum MAE deviates from the expected minimum (in this case, 100:100), the process-based crop growth model could be adjusted in this direction or, in other words, complemented by the knowledge gained from the image time series.
Note that other dynamic growth-influencing variables, like climatic conditions, can be used instead of process-based time-varying biomass, which could lead to even more feasible simulations.

\subsection{Transferability to new site}
With a transferability experiment on the MixedCrop experiment, we aim to investigate the accuracy drop with which the model trained for Mixed-CKA, which takes time (t) as input condition, can be applied to the Mixed-WG site.
The basic requirements are given by the same image size, resolution, crop species, and treatments (see \secref{sec:data}). 
However, this attempt to transfer the growth behavior of Mixed-CKA to images of Mixed-WG poses three main challenges.
First, the growth behavior of conventionally managed CKA differs substantially from that of organically managed WG, as indicated, for instance, by weed abundance.
Second, the spectral image properties are completely different for each time point, so both sites have their own ``style''.
Third, images were not taken at the same time during the growing season at both locations, resulting in images from WG being not only spatially but also temporally out-of-disturbance (OOD).

\tabref{tab:gen_scores_classic} and \tabref{tab:gen_scores_bm} show the transferability quality measured by all evaluation metrics in the bottom line each.
It can be seen that the results show significantly lower accuracies than the ones produced by models trained and tested on Mixed-CKA.
However, the identity predictions still show a high MS-SSIM of 0.92.

The reason for the less accurate results lies in the first two aforementioned challenges, which lead to the predicted images not being well comparable to the reference images on a quantitative basis.
Since the model only knows the style of CKA, but the reference images are in the style of WG, better scores were not expected.
Putting the focus more on qualitative results, the third challenge of temporal OOD leads to corrupted results when the input image is significantly different from the style of the temporally nearest CKA image but is otherwise reliable, which is demonstrated in Appendix \ref{secA:s-ood}.
It shows both failed predictions and reasonable transfer examples, first for time points for which reference images are available, even if they do not match the reference, and second for the entire growing period.

For future experiments, the style could be added as an additional condition in the image-prediction model, or more generally with domain knowledge in the form of site-dependent context variables that influence not only style but also plant growth itself \cite{leonhardtleveraging}.
While this requires a larger training dataset spanning multiple sites and styles, it will ensure even better transferability and could help to merge multiple plant time series affected by various factors influencing factors into a more generic data-driven crop growth model.

\section{Conclusion}\label{sec:conclusion}
In this work, we have shown the capabilities of multi-conditional growth simulation using three datasets Arabidopsis, GrowliFlower, and MixedCrop.
For this purpose, in the first step, we combined several conditions of different types (discrete, continuous, categorical) in an image prediction model, which is a conditional Wasserstein generative adversarial network (CWGAN), to generate multiple realistic high-quality images over time based on a single input image. 
In the second step of growth estimation, we showed that along with classical GAN image evaluation metrics, plant-specific traits such as projected leaf area or biomass can be derived from the generated images and used for evaluation.
The results for MixedCrop were compared with a dynamic process-based growth model.
Here, the combination of data-driven crop growth models, which strongly incorporate the spatio-temporal above-ground phenotype changes, and a process-based crop growth model, which takes the theoretical plant growth knowledge into account, leads to a better understanding of the crop mixture dynamics.
The experiments show that the dried biomass can be estimated more accurately from predicted images the more growth influencing factors are considered, such as in our case, the field treatment or process-based simulated biomasses.
In particular, the integration of process-based model output into a data-driven crop growth model is shown, which is useful to make crop growth predictions more accessible or even to re-calibrate process-based models.
Adding all available conditions into the image prediction model allows plant trait estimation on predicted (artificial) images with similar accuracy as on real images.

Although the additional variability images show the largest uncertainties at the leaf edges, which is realistic, we see space for improvement in the uncertainty integration for long-term growth predictions. 
Since predictions far in the future lead to significant over-confidence in the image prediction model, the noise-input ratio should be adaptively controlled depending on the growth prediction step.
In addition, the challenge of large spectral differences within an image sequence and between sites (``dataset styles'') should be addressed for better model generalizability.


\section*{Acknowledgements}
We thank Julie Krämer, Christian Dahn, and Nils Müller for comprehensive (image) data acquisition and fruitful discussions, as well as Philippe Simonis, Malte Möller, and Manikandan Somu for data pre-processing and selection in the MixedCrop project.

\section*{Funding}
This work was funded by the Deutsche Forschungsgemeinschaft (DFG, German Research Foundation) under Germany’s Excellence Strategy – EXC 2070 – 390732324 and partly funded by the Deutsche Forschungsgemeinschaft (DFG, German Research Foundation) - SFB 1502/1-2022 - Projektnummer: 450058266“.

\normalsize
\bibliography{bibliography_lukas}

\clearpage
\begin{appendices}
\section{Overview of MixedCrop cultivars}\label{secA:mixed-cultivars}
An overview of the faba bean cultivars and spring wheat entities used in the MixedCrop experiment is given in \tabref{tab:mixed-cultivars}.
\begin{table}[t]
\centering
\begin{tabular}{lll}
\toprule
\multirow{2}{*}{FB (Faba bean)}     & A  & Mallory       \\
                                    & B  & Fanfare       \\
\midrule
\multirow{12}{*}{SW (spring wheat)} & 1  & Lennox        \\
                                    & 2  & Anabel        \\
                                    & 3  & Saludo        \\
                                    & 4  & Jasmund       \\
                                    & 5  & Sorbas        \\
                                    & 6  & Quintus       \\
                                    & 7  & KWS Starlight \\
                                    & 8  & Chamsin       \\
                                    & 9  & Sonett        \\
                                    & 10 & SU Ahab       \\
                                    & 11 & Mix-Group 1   \\
                                    & 12 & Mix-Group 2   \\
\bottomrule
\end{tabular}
\caption{Notation overview of species faba bean (FB) with cultivars A and B and spring wheat (SW) with cultivars 1-10 and two additional mixed groups used in this work.}
\label{tab:mixed-cultivars}

\end{table}

\section{Temporal out-of-distribution predictions}\label{secA:t-ood}

\begin{figure*}[t]
    \centering
    \includegraphics[width=1.0\textwidth]{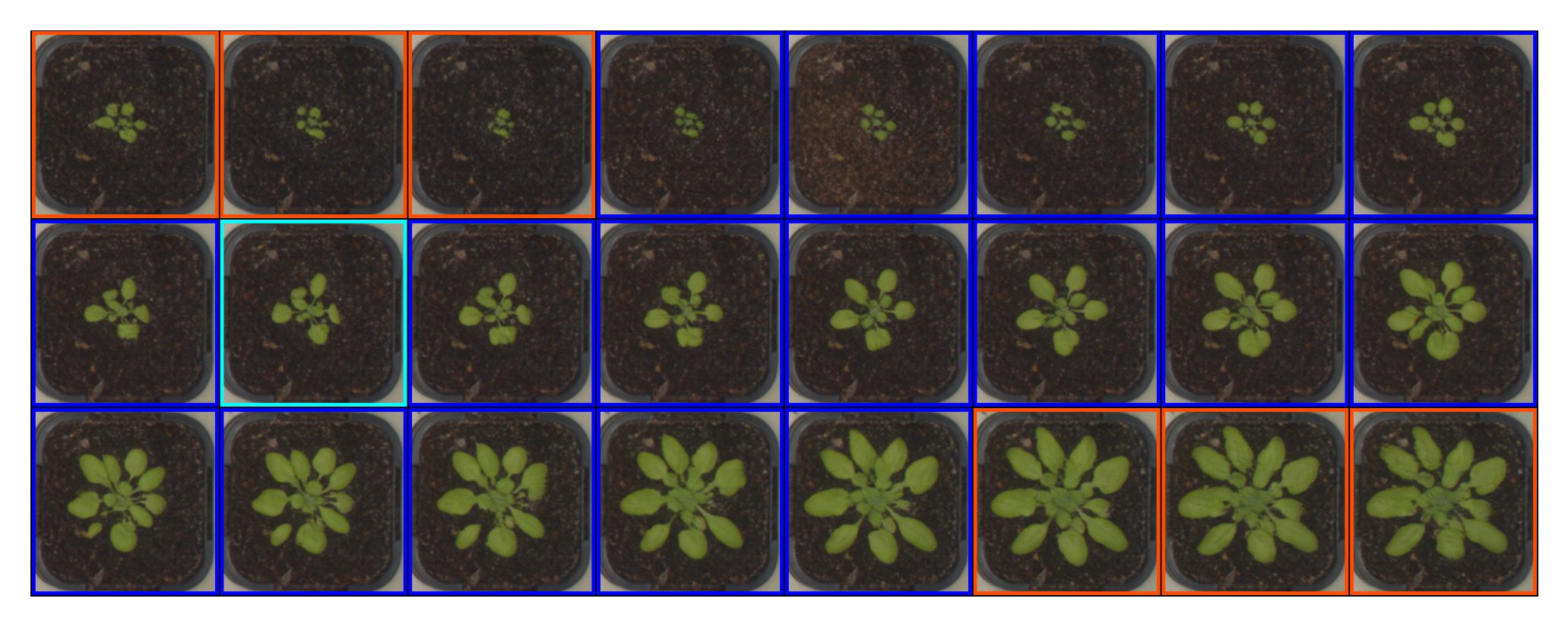}
    \caption{Daily Arabidopsis predictions from \SI{18}{DAS} to \SI{41}{DAS} including temporal OOD images. The input image has a cyan frame, the in-distribution images a blue frame, and the OOD images an orange frame.}
    \label{fig:ood_abd}
\end{figure*}

\begin{figure*}[t]
    \centering
    \includegraphics[width=1.0\textwidth]{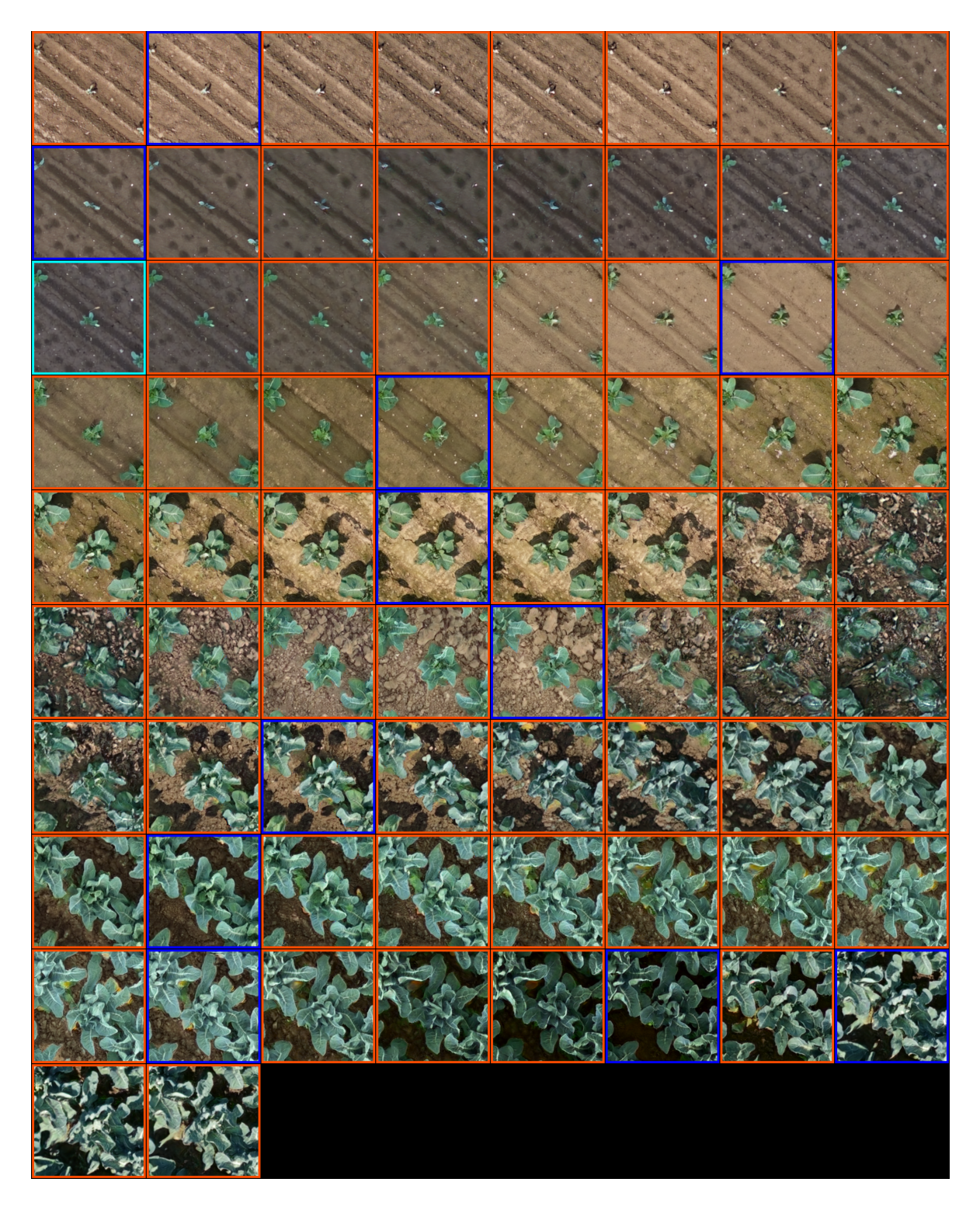}
    \caption{Daily GrowliFlower predictions from \SI{0}{DAP} to \SI{73}{DAP} including temporal OOD images. The input image has a cyan frame, the in-distribution images a blue frame, and the OOD images an orange frame.}
    \label{fig:ood_grf}
\end{figure*}

\begin{figure*}[t]
    \centering
    \includegraphics[width=1.0\textwidth]{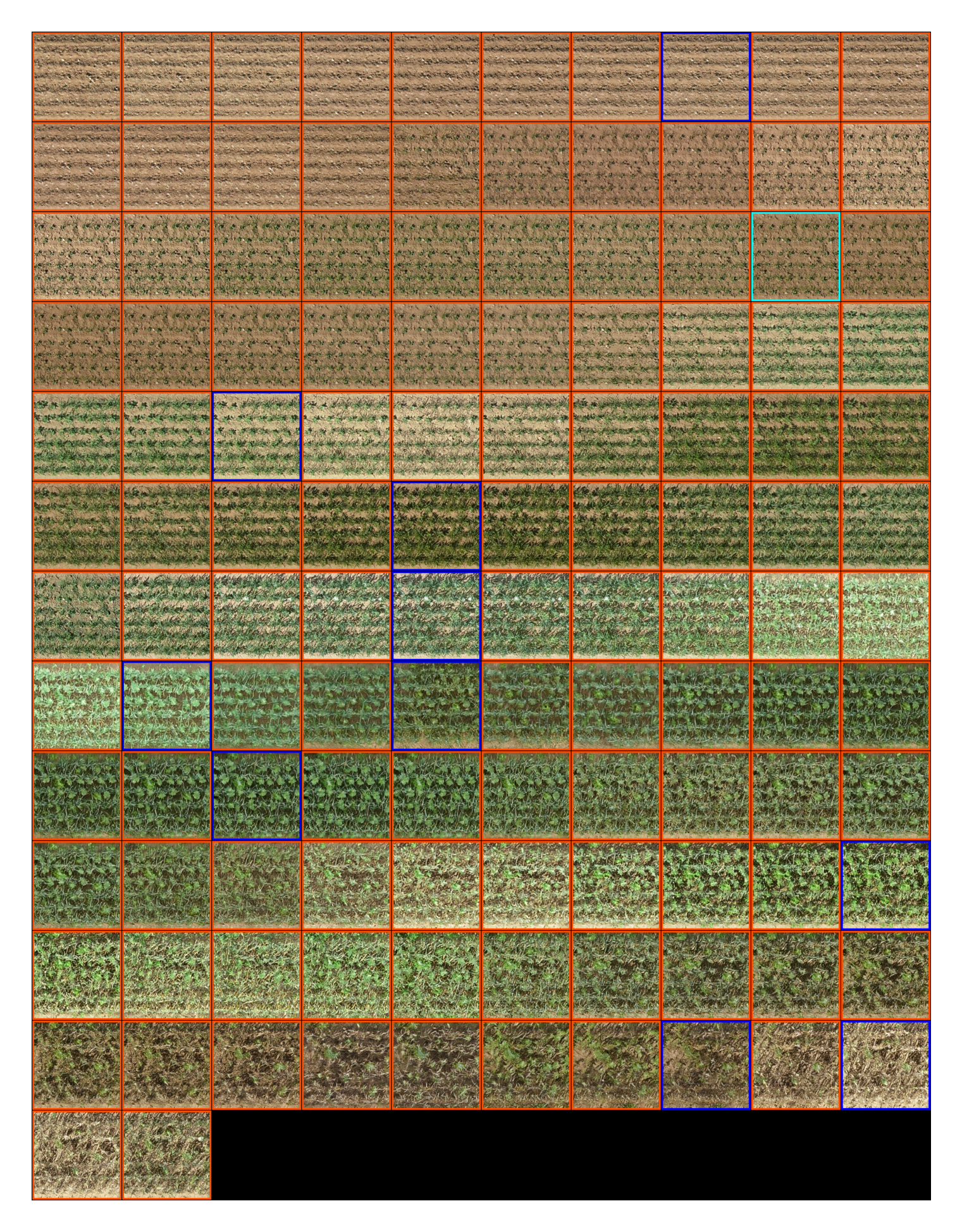}
    \caption{Daily Mixed-CKA predictions from \SI{0}{DAS} to \SI{121}{DAS} including temporal OOD images. The input image has a cyan frame, the in-distribution images a blue frame, and the OOD images an orange frame.}
    \label{fig:ood_mix}
\end{figure*}

By temporal out-of-distribution (OOD), we mean images of growth stages that do not exist in the training dataset.
We use the models from the respective dataset trained only on input image and time as conditions and keep the input image and the noise constant for the visualizations from the entire growth period.
So we iterate over time and generate interpolations if the newly generated image lies between two training images and extrapolations if it lies temporally off the training period (early and late growth phases).
The time increases by one day per image from top left to bottom right. 
The input image has a cyan frame, the in-distribution images a blue frame, and the OOD images an orange frame.
While difficult to evaluate quantitatively because no reference images are available, consistency in the time series is readily apparent for interpolations.
For extrapolations, most predictions are also realistic since plants continue to grow in the short-term extrapolated future.
Notably, interpolation and extrapolation work for Mixed-WG, although the growth stage of the input image is temporally out-of-distribution. 
However, there are exceptions, e.g., the early growth stages of Arabidopsis are too large, and in the third row of Mixed-CKA and Mixed-WG canopy appears and vanishes again.

\section{Spatial out-of-distribution predictions}\label{secA:s-ood}
\begin{figure*}[t]
    \centering
    \includegraphics[width=1.0\textwidth]{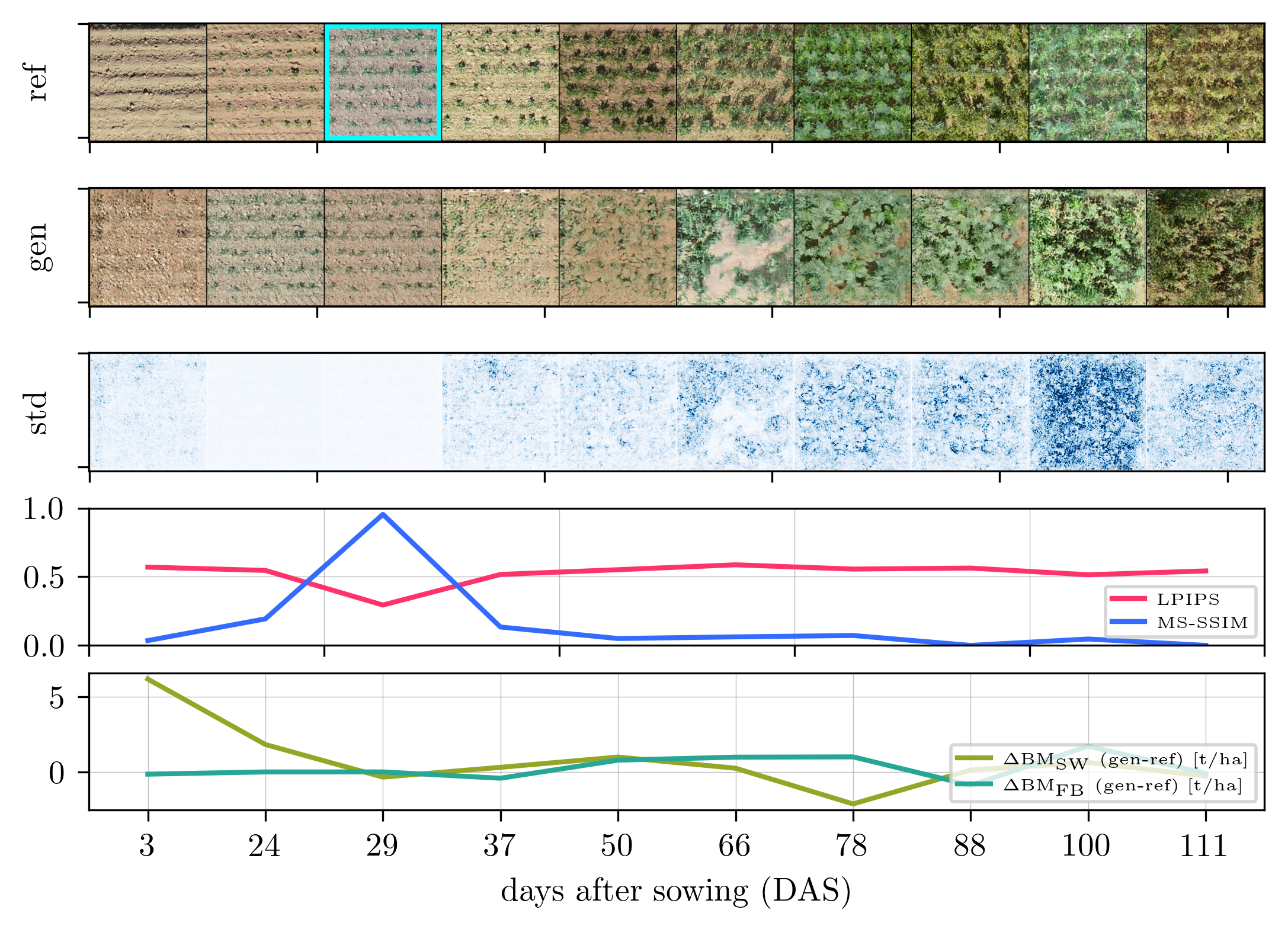}
    \caption{Transferability fails with predictions for Mixed-WG caused by input image \SI{29}{DAS} lying spectrally too far out of the training distribution (Mixed-CKA images).}
    \label{fig:mix-wg_transfer_issue}
\end{figure*}

\begin{figure*}[t]
    \centering
    \includegraphics[width=1.0\textwidth]{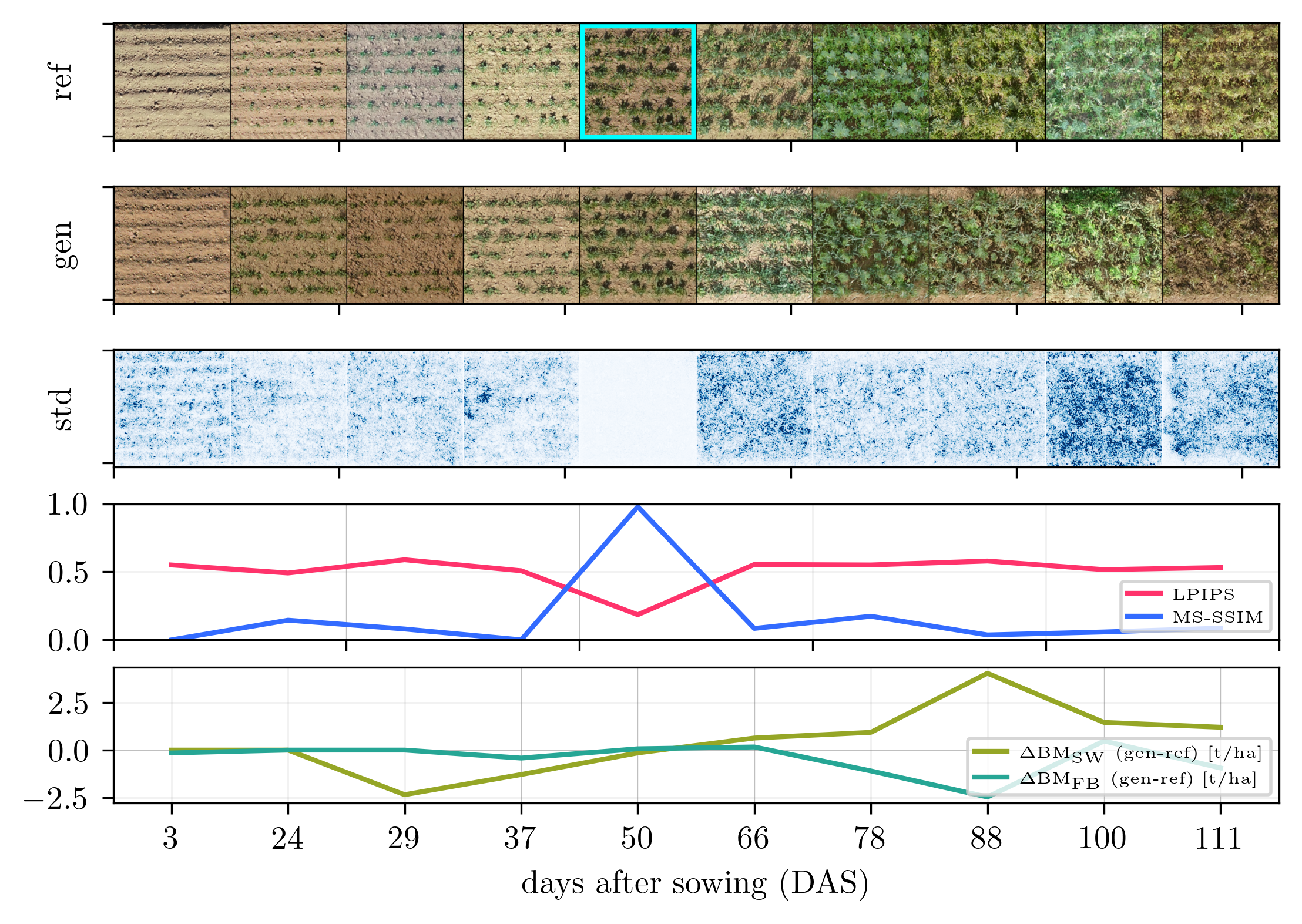}
    \caption{Transferability with prediction results for Mixed-WG input image \SI{50}{DAS} lying spectrally less far away from the \SI{54}{DAS}-images of the training distribution (Mixed-CKA). The predicted images are qualitatively appealing, while they do not compare well with the reference, because the crops of Mixed-CKA and Mixed-WG have different growth patterns.}
    \label{fig:mix-wg_transfer}
\end{figure*}

\begin{figure*}[t]
    \centering
    \includegraphics[width=1.0\textwidth]{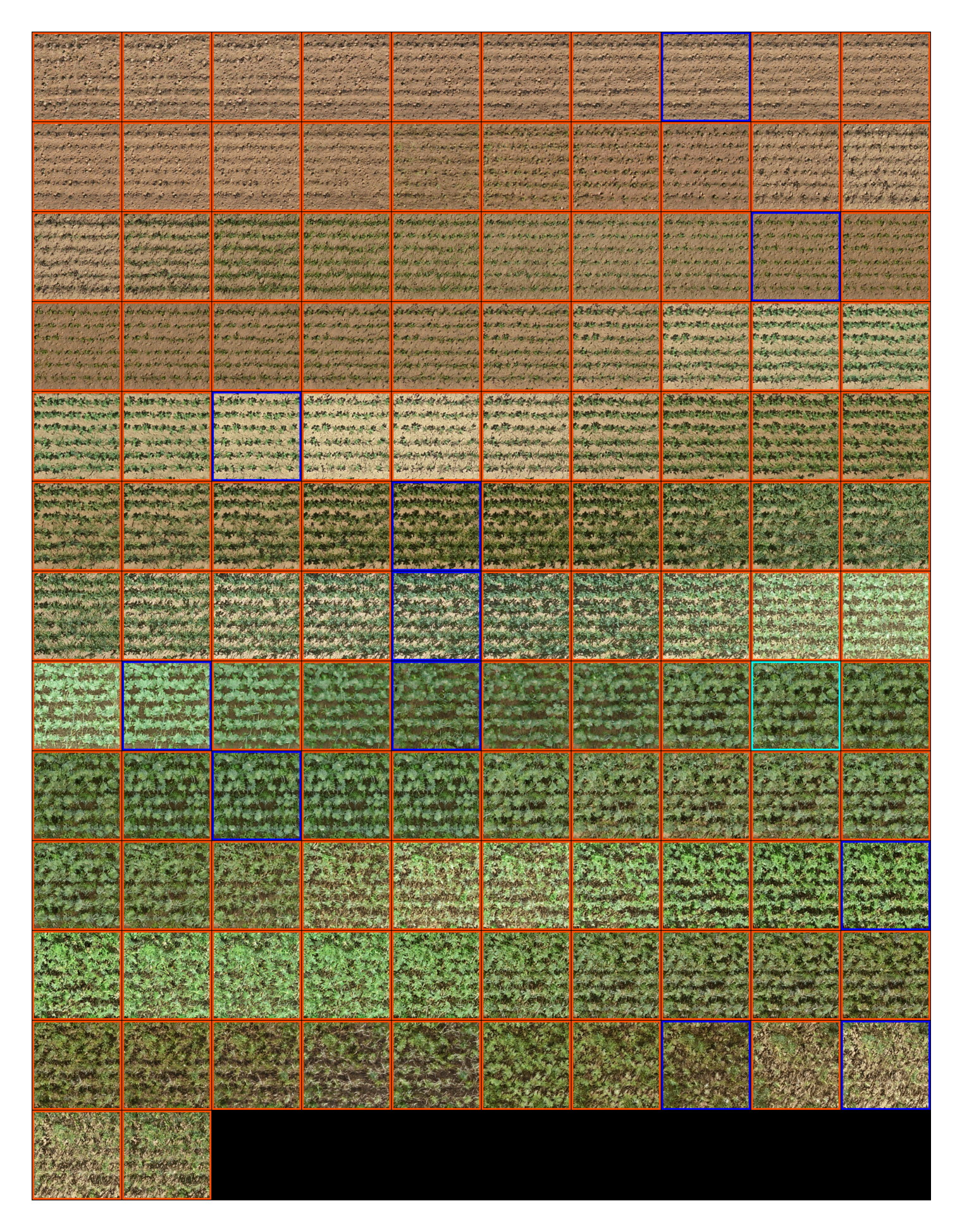}
    \caption{Daily Mixed-WG predictions from \SI{0}{DAS} to \SI{121}{DAS} including temporal OOD images from an image prediction model trained exclusively on Mixed-CKA images.}
    \label{fig:ood_mix-wg}
\end{figure*}

If the image prediction model is applied to a site on whose plant image time series the model has not been trained, it is spatially out-of-distribution.
Additionally, it is likely that the dataset is temporally OOD if the drone flyovers did not occur on the same days of the growing season, and thus, the time of the new input image does not exist in the training data.
There, transferability fails when the spectral differences between the test image and the nearby time points in the training dataset are too large, such as \SI{29}{DAS} of Mixed-WG, as \figref{fig:mix-wg_transfer_issue} illustrates.
Some of the predicted images become blurry, and holes appear in the crop rows, which also causes the biomass estimation to give unreliable, non-usable results.
Likewise, \figref{fig:mix-wg_transfer} demonstrates that the model can produce reasonable results despite spatio-temporal OOD, where, compared to \figref{fig:mix-wg_transfer_issue}, the same field patch but a different input image (21 days later) is used.
Realistic results can also be achieved when not only the input image but also the images to be predicted are temporally and spatially OOD, as depicted in \figref{fig:ood_mix-wg}.

\end{appendices}
\end{document}